\documentclass[10pt,twocolumn,letterpaper]{article}

\usepackage{iccv}              
\usepackage{multirow}
\usepackage{colortbl}
\usepackage[ruled,vlined]{algorithm2e}
\definecolor{commentcolor}{RGB}{110,154,155}   
\usepackage{listings}
\definecolor{iccvblue}{rgb}{0.21,0.49,0.74}
\usepackage[pagebackref,breaklinks,colorlinks,allcolors=iccvblue]{hyperref}

\def\paperID{6403} 
\def\confName{ICCV}
\def\confYear{2025}

\title{SVTRv2: CTC Beats Encoder-Decoder Models in Scene Text Recognition}

\author{
Yongkun Du$^{1}$, Zhineng Chen$^{1}\thanks{Corresponding Author}$, Hongtao Xie$^{2}$, Caiyan Jia$^{3}$, Yu-Gang Jiang$^{1}$ \\
$^1$Institute of Trustworthy Embodied AI, Fudan University, China\\
$^2$School of Information Science and Technology, USTC, China \\
$^3$School of Computer Science and Technology, Beijing Jiaotong University, China\\
{\tt\small
ykdu23@m.fudan.edu.cn,
\{zhinchen, ygj\}@fudan.edu.cn,
htxie@ustc.edu.cn,
cyjia@bjtu.edu.cn}
}

\begin{document}
\maketitle

\begin{abstract}

Connectionist temporal classification (CTC)-based scene text recognition (STR) methods,  e.g., SVTR, are widely employed in OCR applications, mainly due to their simple architecture, which only contains a visual model and a CTC-aligned linear classifier, and therefore fast inference. However, they generally exhibit worse accuracy than encoder-decoder-based methods (EDTRs) due to struggling with text irregularity and linguistic missing. To address these challenges, we propose SVTRv2, a CTC model endowed with the ability to handle text irregularities and model linguistic context. First, a multi-size resizing strategy is proposed to resize text instances to appropriate predefined sizes, effectively avoiding severe text distortion. Meanwhile, we introduce a feature rearrangement module to ensure that visual features accommodate the requirement of CTC, thus alleviating the alignment puzzle. Second, we propose a semantic guidance module. It integrates linguistic context into the visual features, allowing CTC model to leverage language information for accuracy improvement. This module can be omitted at the inference stage and would not increase the time cost. We extensively evaluate SVTRv2 in both standard and recent challenging benchmarks, where SVTRv2 is fairly compared to popular STR models across multiple scenarios, including different types of text irregularity, languages, long text, and whether employing pretraining. SVTRv2 surpasses most EDTRs across the scenarios in terms of accuracy and inference speed. Code: \url{https://github.com/Topdu/OpenOCR}.

\end{abstract}

\section{Introduction}
\label{sec:intro}

\begin{figure}
  \centering
\includegraphics[width=0.48\textwidth]{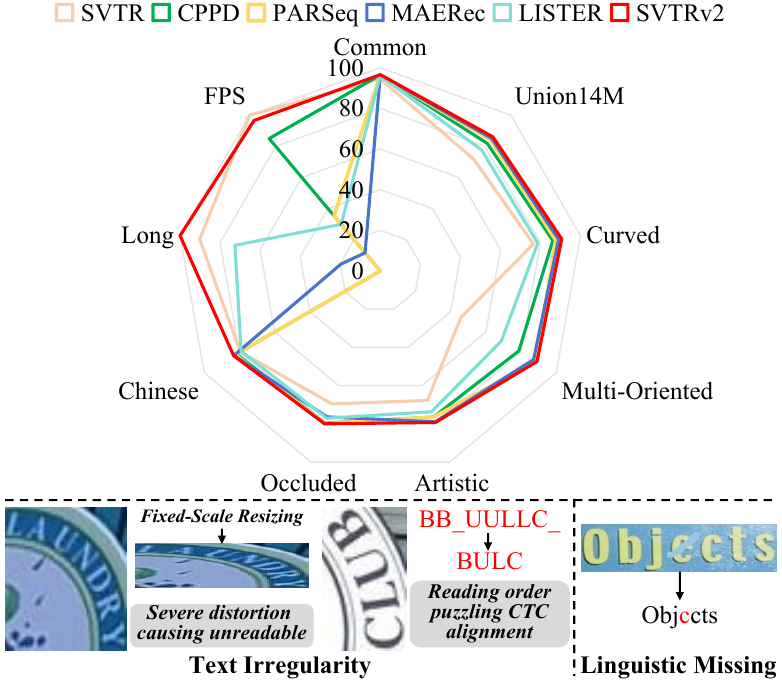} 
  \caption{\textbf{Top}: comparison with previous methods~\cite{duijcai2022svtr,du2023cppd,BautistaA22PARSeq,jiang2023revisiting,iccv2023lister} best in a single scenario, where long text recognition accuracy (Long) and FPS are normalized. Our SVTRv2 achieves the new state of the arts in every scenario except for FPS. Nevertheless, SVTRv2 is still the fastest compared to all the EDTRs. \textbf{Bottom}: challenges caused by text irregularity and linguistic missing.}
  \label{fig:fig1}
\end{figure}

As a task of extracting text from natural images, scene text recognition (STR) has garnered considerable interest over decades. Unlike text from scanned documents, scene text often exists within complex natural scenarios, posing challenges such as background noise, text distortions, irregular layouts, artistic fonts~\cite{ChenJZLW21str_survey}, etc. To tackle these challenges, a variety of STR methods have been developed and they can be roughly divided into two categories, i.e., connectionist temporal classification (CTC)-based methods and encoder-decoder-based methods (EDTRs).

Typically, CTC-based methods~\cite{shi2017crnn,duijcai2022svtr,hu2020gtc,ppocrv3} employ a single visual model to extract image features and then apply a CTC-aligned linear classifier~\cite{CTC} to predict recognition results. This straightforward architecture provides advantages such as fast inference, which makes them especially popular in OCR applications. However, these models struggle to handle text irregularity, i.e., text distortions, varying layouts, etc. As a consequence, attention-based decoders are introduced as alternatives, leading to a series of EDTRs~\cite{shi2019aster,Sheng2019nrtr,pr2019MORAN,li2019sar,wang2020aaai_dan,yu2020srn,cvpr2020seed,zhang2020autostr,yue2020robustscanner,fang2021abinet,Wang_2021_visionlan,wang2022tip_PETR,BautistaA22PARSeq,mgpstr,du2023cppd,ijcai2023LPV,iccv2023lister,zheng2024cdistnet,TPAMI2022ABINetPP,yang2024class_cam,Wei_2024_busnet,Xu_2024_CVPR_OTE,levocr,xie2022toward_cornertrans,Guan_2023_CVPR_SIGA,Guan_2023_ICCV_CCD,zhou2024cff,Zhao_2024_CVPR_E2STR,zhong_2024_acmmm_vlreader,zhao_2025_tip_clip4str,zhao_2024_acmmm_dptr}. These methods exhibit superior performance in complex scenarios by leveraging multi-modal cues, including visual~\cite{Wang_2021_visionlan,ijcai2023LPV,du2023cppd,Xu_2024_CVPR_OTE}, linguistic~\cite{Sheng2019nrtr,yu2020srn,fang2021abinet,cvpr2020seed}, and positional~\cite{yue2020robustscanner,zheng2024cdistnet,iccv2023lister} ones, which are largely missed in current CTC models. As depicted in the top of Fig.~\ref{fig:fig1}, compared to SVTR~\cite{duijcai2022svtr}, a leading CTC model adopted by famous commercial OCR engines~\cite{ppocrv3}, EDTRs achieve superior results in scenarios \cite{Wang_2021_visionlan,jiang2023revisiting,chen2021benchmarking} such as curved, multi-oriented, artistic, occluded, and Chinese text.

The inferior accuracy of CTC models can be attributed to two primary factors. First, these models struggle with irregular text, as CTC alignment presumes that the text appears in a near canonical left-to-right order~\cite{ChenJZLW21str_survey,whatwrong}, which is not always true, particularly in complex scenarios. Second, CTC models seldom encode linguistic information, which is typically accomplished by the decoder of EDTRs. While recent advancements deal with the two issues by employing text rectification~\cite{shi2019aster,pr2019MORAN,zheng2023tps++}, developing 2D CTC~\cite{arxiv2019_2dctc}, utilizing masked image modeling~\cite{Wang_2021_visionlan,ijcai2023LPV}, etc., the accuracy gap between CTC and EDTRs remains significant, indicating that novel solutions still need to be investigated.

In this paper, our aim is to build more powerful CTC models by better handling text irregularity and integrating linguistic context. For the former, we address this challenge by first extracting discriminative features and then better aligning them. First, existing methods uniformly resize text images with various shapes to a fixed size before feeding into the visual model. We question the rationality of this resizing, which easily causes unnecessary text distortion, making the text difficult to read, as shown in the bottom-left of Fig.~\ref{fig:fig1}. To this end, a multi-size resizing (MSR) strategy is proposed to resize the text instance to a proper predefined size based on its aspect ratio, thus minimizing text distortion and ensuring the discrimination of the extracted visual features. Second, irregular text may be rotated significantly, and the character arrangement does not align with the reading order of the text, causing the puzzle of CTC alignment, as shown in the bottom-center example in Fig.~\ref{fig:fig1}. To solve this, we introduce a feature rearrangement module (FRM). It rearranges visual features with first a horizontal rearrangement and then a vertical rearrangement to identify and prioritize relevant features. FRM maps 2D visual features into a sequence aligned with the text's reading order, thus effectively alleviating the alignment puzzle. Consequently, CTC models integrating MSR and FRM can recognize irregular text well, without using rectification modules or attention-based decoders.

As for the latter, the mistakenly recognized example shown in the bottom-right of Fig.~\ref{fig:fig1} clearly highlights the necessity of integrating linguistic information. Since CTC models directly classify visual features, we have to endow the visual model with linguistic context modeling capability, which is less discussed previously. Inspired by guided training of CTC (GTC)~\cite{hu2020gtc,ppocrv3} and string matching-based recognition~\cite{du2024smtr}, we propose a semantic guidance module (SGM), a new scheme that solely leverages surrounding string context to model the target character. This approach effectively guides the visual model in capturing linguistic context. During inference, SGM can be omitted and would not increase the time cost.

With these contributions, we develop SVTRv2, a novel CTC-based method whose recognition ability has been largely enhanced, while still maintaining a simple inference architecture and fast speed. To thoroughly validate SVTRv2, we conducted extensive and comparative experiments on benchmarks including standard regular and irregular text~\cite{whatwrong}, Union14M-Benchmark~\cite{jiang2023revisiting}, occluded scene text~\cite{Wang_2021_visionlan}, long text~\cite{du2024smtr}, and Chinese text~\cite{chen2021benchmarking}. The results demonstrate that SVTRv2 consistently outperforms all the compared EDTRs across the evaluated scenarios in terms of accuracy and speed. Moreover, a simple pretraining on SVTRv2 yields highly competitive accuracy compared to the pretraining-based EDTRs advances \cite{Zhao_2024_CVPR_E2STR,zhong_2024_acmmm_vlreader,zhao_2025_tip_clip4str,zhao_2024_acmmm_dptr}, highlighting its effectiveness and broad applicability.

In addition, recent advances~\cite{jiang2023revisiting,Rang_2024_CVPR_clip4str} indicated the importance of large-scale real-world datasets in improving STR performance. However, many STR models primarily derived from synthetic data~\cite{Synthetic,jaderberg14synthetic}, which fail to fully represent real-world complexities and lead to performance limitations, particularly on challenging scenarios. Meanwhile, we observe that existing large-scale real-word training datasets~\cite{BautistaA22PARSeq,jiang2023revisiting,Rang_2024_CVPR_clip4str} overlap with Union14M-Benchmark, causing a small overlapping between training and test data, thus the results reported in~\cite{jiang2023revisiting} should be updated. As a result, we introduce \textit{U14M-Filter}, a rigorously filtered version of the real-world training dataset \textit{Union14M-L}~\cite{jiang2023revisiting}. Then, we systematically reproduced and retrained 24 mainstream STR methods from scratch based on \textit{U14M-Filter}. These methods are thoroughly evaluated on Union14M-Benchmark. Their accuracy, model size, and inference time constitute a comprehensive and reliable new benchmark for future reference.

\begin{figure*}[t]
  \centering
\includegraphics[width=0.98\textwidth]{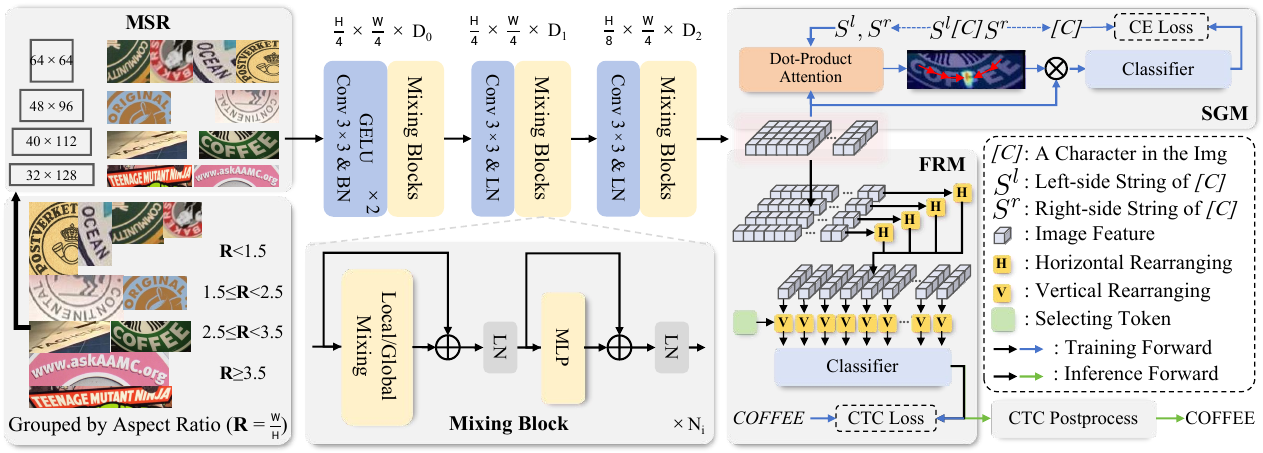} 
  \caption{An illustrative overview of SVTRv2. The text is first resized according to multi-size resizing (MSR), then experiences feature extraction. During training both the semantic guidance module (SGM) and feature rearrangement module (FRM) are employed, which are responsible for linguistic context modeling and CTC-oriented feature rearrangement, respectively. Only FRM is retained during inference.}
  \label{fig:svtrv2_overview}
\end{figure*}

\section{Related Work}

\noindent\textbf{Irregular text recognition}~\cite{Risnumawan2014cute,SVTP,jiang2023revisiting} has posed a significant challenge in STR due to the diverse variation of text instances, where CTC-based methods~\cite{shi2017crnn,duijcai2022svtr,hu2020gtc,ppocrv3} are often less effective. To address this, some methods~\cite{shi2019aster,cvpr2020seed,zhang2020autostr,zheng2024cdistnet,yang2024class_cam,duijcai2022svtr,zheng2023tps++} incorporate rectification modules~\cite{shi2019aster,pr2019MORAN,zheng2023tps++} that aim to transform irregular text into more regular format. While more methods utilize attention-based decoders~\cite{wang2020aaai_dan,li2019sar,Sheng2019nrtr,yue2020robustscanner,du2023cppd,du2024smtr}, which employ the attention mechanism to dynamically localize characters regardless of text layout, and thus are less affected. 
However, these methods generally have tailored training hyper-parameters. For example, rectification modules~\cite{shi2019aster,pr2019MORAN,zheng2023tps++} typically specify a fixed output image size (e.g. 32$\times$128), which is not always a suitable choice. While attention-based decoders~\cite{wang2020aaai_dan,li2019sar,Sheng2019nrtr,yue2020robustscanner,du2023cppd,du2024smtr} generally set the maximum recognition length to 25 characters, therefore, longer text cannot be recognized, as shown in Fig.~\ref{fig:case_long}.

\noindent\textbf{Linguistic context modeling}. There are several ways of modeling linguistic context. One major branch is auto-regressive (AR)-based STR methods~\cite{shi2019aster,wang2020aaai_dan,Sheng2019nrtr,li2019sar,jiang2023revisiting,xie2022toward_cornertrans,zheng2024cdistnet,Xu_2024_CVPR_OTE,yang2024class_cam,zhou2024cff,du2024igtr}, which utilize previously decoded characters to model contextual cues. However, their inference speed is slow due to the character-by-character decoding nature. Some other methods~\cite{yu2020srn,fang2021abinet,MATRN,BautistaA22PARSeq} integrate external language models to model linguistic context and correct the recognition results. While effective, the linguistic context is purely text-based, making it challenging to adapt them to the visual model of CTC models. There are also some studies~\cite{cvpr2020seed,Wang_2021_visionlan,ijcai2023LPV} to model linguistic context with visual information only using pretraining based on masked image modeling~\cite{HeCXLDG22_mae,Bao0PW22_beit}. However, they still depend on attention-based decoders to utilize linguistic information, not integrating linguistic cues into the visual model, thus limiting their effectiveness in enhancing CTC models.

\section{Method}

Fig.~\ref{fig:svtrv2_overview} illustrates the overview of SVTRv2. A text image is first resized by MSR to the closest aspect ratio, forming the input  \( \mathbf{X} \in \mathbb{R}^{3 \times H \times W} \), which then experiences three consecutive feature extraction stages, yielding visual features $\mathbf{F}\in \mathbb{R}^{\frac{H}{8} \times \frac{W}{4} \times D_2}$. During training, $\mathbf{F}$ is fed into both SGM and FRM. SGM guides SVTRv2 to model linguistic context, while FRM rearranges $F$ into the character feature sequence $\mathbf{\tilde{F}} \in \mathbb{R}^{\frac{W}{4} \times D_2}$, which is synchronized with the text reading order and aligns with the label sequence. During inference, the SGM is discarded for efficiency.

\subsection{Multi-Size Resizing (MSR)}

Previous works typically resize irregular text images to a fixed size, such as $32 \times 128$, which may cause undesired text distortion and severely affect the quality of extracted visual features. To address this issue, we propose a simple yet effective multi-size resizing (MSR) strategy that resizes text shapes based on the aspect ratio ($R=\frac{W}{H}$). Specifically, we define four specific sizes: [64, 64], [48, 96], [40, 112], and [32, $\lfloor 
R\rfloor \times$ 32], respectively corresponding to aspect ratio: $R<$ 1.5 ($R_1$), 1.5 $\leq R <$ 2.5 ($R_2$), 2.5 $\leq R <$ 3.5 ($R_3$), and $R\geq$ 3.5 ($R_4$). Note that the first three buckets are fixed thus text instances in the same one can be trained in batch, while the fourth one can handle long text without introducing significant distortion.
Therefore, MSR allows text instances adaptively resized under the principles of roughly maintaining their aspect ratios, and significant text distortion caused by resizing is almost eliminated.

\subsection{Visual Feature Extraction}

Motivated by SVTR~\cite{duijcai2022svtr}, the visual model of SVTRv2 comprises three stages, with \( \text{stage}_i \) containing \( N_i \) mixing blocks, as illustrated in Fig.~\ref{fig:svtrv2_overview}. To extract discriminative visual features, we devise two types of mixing blocks: local and global. Unlike SVTR, for being able to handle multiple sizes, we do not use absolute positional encoding. In contrast, to model positional information, we implement local mixing as two consecutive grouped convolutions alternative to window attention \cite{duijcai2022svtr}, and effectively capturing local character features, such as edges, textures, and strokes. Meanwhile, global mixing is realized by the multi-head self-attention (MHSA) mechanism~\cite{NIPS2017_attn}. This mechanism performs global contextual modeling on features, thereby enhancing the model's comprehension of inter-character relationships and the overall text image. Both the number of groups in the grouped convolution and the number of heads in MHSA are set to \( \frac{D_i}{32} \). Similar to SVTR~\cite{duijcai2022svtr}, by adjusting hyper-parameters $N_i$ and $D_i$, we derive three variants of SVTRv2 with different capacities, i.e., Tiny (T), Small (S), and Base (B), which are detailed in \textit{Suppl. Sec.}~7.

\subsection{Feature Rearranging Module (FRM)}

To address the alignment puzzle caused by text irregularities like rotated text, we propose a feature rearrangement module (FRM). This module rearranges the visual features \( \mathbf{F} \in \mathbb{R}^{(\frac{H}{8} \times \frac{W}{4}) \times D_2} \) into a sequence \( \mathbf{\tilde{F}} \in \mathbb{R}^{\frac{W}{4} \times D_2} \) that conforms to the CTC alignment requirement. We model the rearrangement process as a soft mapping using a probability matrix \( \mathbf{M} \in \mathbb{R}^{\frac{W}{4} \times (\frac{H}{8} \times \frac{W}{4})} \), where each element \( \mathbf{M}_{m,i,j} \) represents the probability of mapping the original feature \( \mathbf{F}_{i,j} \) to the rearranged feature \( \mathbf{\tilde{F}}_m \), where \( i \in \{1, 2, \ldots, \frac{H}{8}\} \) and \( j, m \in \{1, 2, \ldots, \frac{W}{4}\} \). Consequently, the softly rearrangement process is formalized as: \(\mathbf{\tilde{F}} = \mathbf{M} \times \mathbf{F}.
\)

FRM is responsible for learning the matrix \( \mathbf{M} \). To ensure that the module is sensitive to text orientation, we decompose the learning process of \( \mathbf{M} \) into two sequential steps: horizontal rearranging and vertical rearranging. As illustrated in Fig.~\ref{fig:svtrv2_overview}, the horizontal one processes each row of the visual feature \( \mathbf{F} \), denoted as \( \mathbf{F}_i \in \mathbb{R}^{\frac{W}{4} \times D_2} \), to learn a horizontal rearrangement matrix \( \mathbf{M}^h_i \in \mathbb{R}^{\frac{W}{4} \times \frac{W}{4}} \), which rearranges visual features along the horizontal direction. We implement this process using a multi-head self-attention mechanism as follows:  
\begin{gather}
\label{eq:h_matrix}
\mathbf{M}^h_i = \sigma\left(\mathbf{F}_i\mathbf{W}^q_i\left(\mathbf{F}_i\mathbf{W}^k_i\right)^t\right) \\ \notag
\mathbf{F}^{h'}_i = \text{LN}(\mathbf{M}^h_i\mathbf{F}_i\mathbf{W}^v_i+\mathbf{F}_i), \mathbf{F}^h_i = \text{LN}(\text{MLP}(\mathbf{F}^{h'}_i)+\mathbf{F}^{h'}_i)
\end{gather}
where \( \mathbf{W}^q_i, \mathbf{W}^k_i, \mathbf{W}^v_i \in \mathbb{R}^{D_2 \times D_2} \) are learnable weights, \( \sigma \) and  \( ()^t \) denote the softmax function and matrix transpose operation, respectively. LN ans MLP means Layer Normalization and Multi-Layer Perceptron with an expansion rate of 4, respectively. \( \mathbf{F}^h = \{\mathbf{F}^h_1,\mathbf{F}^h_2,\ldots,\mathbf{F}^h_{\frac{H}{8}}\} \) represents the horizontally rearranged visual features.

Similarly, the vertical rearrangement processes visual features column-wise. Unlike the horizontal step, we introduce a selecting token \( \mathbf{T}^s \), which interacts with all column features via cross-attention to learn a vertical rearrangement matrix \( \mathbf{M}^v \). The elements of \( \mathbf{M}^v \) represent the probability of mapping column features to the rearranged features, yielding \( \mathbf{\tilde{F}} \) as follows:  
\begin{equation}
\label{eq:v_matrix}
\mathbf{M}^v_j = \sigma\left(\mathbf{T}^s\left(\mathbf{F}^h_{:,j}\mathbf{W}^k_j\right)^t\right),~ \mathbf{F}^v_j = \mathbf{M}^v_j\mathbf{F}^h_{:,j}\mathbf{W}^v_j
\end{equation}
where \( \mathbf{W}^q_j, \mathbf{W}^k_j, \mathbf{W}^v_j \in \mathbb{R}^{D_2 \times D_2} \) are learnable weights.

We denote \( \mathbf{F}^v = \{\mathbf{F}^v_1,\mathbf{F}^v_2,\ldots,\mathbf{F}^v_{\frac{W}{4}}\} \in \mathbb{R}^{\frac{W}{4} \times D_2} \) as the final rearranged feature sequence \( \mathbf{\tilde{F}} \). The predicted character sequence \( \mathbf{\tilde{Y}}_{ctc} \in \mathbb{R}^{\frac{W}{4} \times N_c} \) is then obtained by passing \( \mathbf{\tilde{F}} \) through the classifier \(\mathbf{\tilde{Y}}_{ctc} = \mathbf{\tilde{F}} \mathbf{W}^{ctc}, \)
where \( \mathbf{W}^{ctc} \in \mathbb{R}^{D_2 \times N_c} \) is the learnable weight matrix. The predicted sequence is then aligned with the ground truth sequence \( \mathbf{Y} \) using the CTC alignment rule.

\subsection{Semantic Guidance Module (SGM)}

\begin{figure}[t]
  \centering
\includegraphics[width=0.48\textwidth]{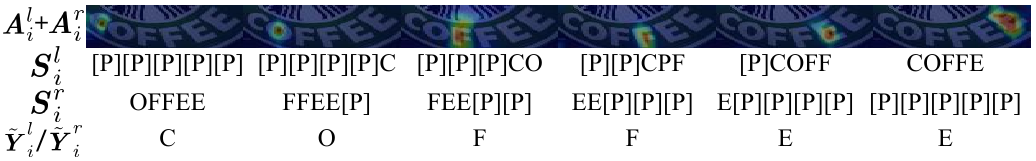} 
\caption{Visualization of attention maps when recognizing the target character by string matching on both sides, where $\textit{l}_i$ is set to 5. [P] denotes the padding symbol.}
\label{fig:smtr}
\end{figure}

CTC models classify visual features directly to obtain recognition results. This scheme inherently requires that the linguistic context must be incorporated into visual features, only that the CTC could benefit from it. In light of this, we propose a semantic guidance module (SGM) as follows.

For a text image with character labels \( \mathbf{Y}\) = \{\(c_1\), \(c_2\), \(\dots\), \(c_L\)\}, where \(c_i \) is the $i$-th character, we define its contextual information as the surrounding left string \( \mathbf{S}^l_i \) = \( \{c_{i-l_s}, \dots, c_{i-1}\} \) and right string \( \mathbf{S}^r_i\) = \(\{c_{i+1}, \dots, c_{i+l_s}\} \), where \( l_s \) denotes the size of the context window. SGM's role is to guide the visual model to integrate context from both \( \mathbf{S}^l_i \) and \( \mathbf{S}^r_i \) into visual features.

We describe the process using the left string \( \mathbf{S}^l_i \). First, the characters in \( \mathbf{S}^l_i \) are mapped to string embeddings \( \mathbf{E}^{l}_i \in \mathbb{R}^{l_s \times D_2} \). Then, these embeddings are encoded to create a hidden representation \( \mathbf{Q}^l_i \in \mathbb{R}^{1 \times D_2} \), representing the context of the left-side string \( \mathbf{S}^l_i \). In the following, the attention map \( \mathbf{A}^l_i \) is computed by applying a dot product between the hidden representation \( \mathbf{Q}^l_i \) and the visual features \( \mathbf{F} \), transformed by learned weight matrices \( \mathbf{W}^q \) and \( \mathbf{W}^k \). The formulation is as follows:
\begin{gather}
    \mathbf{Q}^l_i = \text{LN}\left(\sigma\left(\mathbf{T}^l \mathbf{W}^q \left(\mathbf{E}^{l}_i \mathbf{W}^k \right)^t\right)\mathbf{E}^{l}_i \mathbf{W}^v + \mathbf{T}^l\right) \\ \notag
    \mathbf{A}^l_i = \sigma\left(\mathbf{Q}^l_i \mathbf{W}^q \left(\mathbf{F} \mathbf{W}^k\right)^t \right),~ \mathbf{F}^l_i = \mathbf{A}^l_i \mathbf{F} \mathbf{W}^v
\end{gather}
where \( \mathbf{T}^l \in \mathbb{R}^{1 \times D_2} \) represents a predefined token encoding the left-side string. The attention map \( \mathbf{A}^l_i \) is used to weight the visual features \( \mathbf{F} \), producing a feature \( \mathbf{F}^l_i \in \mathbb{R}^{1 \times D_2} \) corresponding to character \( c_i \). After processing through the classifier $\mathbf{\tilde{Y}}^l_i=\mathbf{F}^l_i \mathbf{W}^{sgm}$, the predicted class probabilities \( \mathbf{\tilde{Y}}^l_i \in \mathbb{R}^{1 \times N_c} \) for \( c_i \) is obtained to calculate the cross-entropy loss, where $\mathbf{W}^{sgm} \in \mathbb{R}^{D_2 \times N_c}$ is learnable weights. 

The weight of the attention map \( \mathbf{A}^l_i \) records the relevance of \( \mathbf{Q}^l_i \) to visual features \( \mathbf{F} \), and moreover, \( \mathbf{Q}^l_i \) represents the context of string \( \mathbf{S}^l_i \). So only when the visual model incorporates the context from \( \mathbf{S}^l_i \) into the visual features of the target character \( c_i \), the attention map \( \mathbf{A}^l_i \) can maximize the relevance between \( \mathbf{Q}^l_i \) and visual features of that character, thus accurately highlighting the corresponding position of character \( c_i \), as shown in Fig.~\ref{fig:smtr}. A similar process can be applied to the right-side string \( \mathbf{S}^r_i \), where the corresponding attention map \( \mathbf{A}^r_i \) and visual feature \( \mathbf{F}^r_i \) contribute to the prediction \( \mathbf{\tilde{Y}}^r_i \). By leveraging the above scheme during training, SGM effectively guides the visual model in integrating linguistic context into visual features. Consequently, even when SGM is not used during inference, the linguistic context can still be maintained and enhancing the accuracy of CTC models. 

Note that although SGM is a decoder-based module, during inference it has been discarded and SVTRv2 becomes a purely CTC model. In contrast, previous methods, such as VisionLAN~\cite{Wang_2021_visionlan} and LPV~\cite{ijcai2023LPV}, despite modeling linguistic context using visual features, still rely on attention-based decoders to activate linguistic information during inference, a process that is incompatible with CTC models.

\subsection{Optimization Objective}

During training, the optimization objective is to minimize the loss $\mathcal{L}$, which comprises $\mathcal{L}_{ctc}$ and $\mathcal{L}_{sgm}$ as listed below:
\begin{align}
    \mathcal{L}_{ctc} &= CTCLoss(\mathbf{\tilde{Y}}_{ctc}, \mathbf{Y}) \\ \notag
    \mathcal{L}_{sgm} &= \frac{1}{2L} \sum\nolimits_{i=1}^{L}(CE(\mathbf{\tilde{Y}}^l_i, c_i) + CE(\mathbf{\tilde{Y}}^r_i, c_i)) \\ \notag
    \mathcal{L} &= \lambda_1 \mathcal{L}_{ctc} + \lambda_2 \mathcal{L}_{sgm} 
\end{align}
\noindent where \(CE\) represents the cross-entropy loss, \(\lambda_1\) and \(\lambda_2\) are weighting parameters setting to 0.1 and 1, respectively.

\section{Experiments}

\subsection{Datasets and Implementation Details}
\label{sec:Implementation}

We evaluate SVTRv2 across multiple benchmarks covering diverse scenarios. They are: 1) six common regular and irregular benchmarks (\textit{Com}), including ICDAR 2013 (\textit{IC13})~\cite{icdar2013}, Street View Text (\textit{SVT})~\cite{Wang2011SVT}, IIIT5K-Words (\textit{IIIT5K})~\cite{IIIT5K}, ICDAR 2015 (\textit{IC15})~\cite{icdar2015}, Street View Text-Perspective (\textit{SVTP})~\cite{SVTP} and \textit{CUTE80}~\cite{Risnumawan2014cute}. For IC13 and IC15, we use the versions with 857 and 1811 images, respectively; 2) the recent Union14M-Benchmark (\textit{U14M})~\cite{jiang2023revisiting}, which includes seven challenging subsets: \textit{Curve}, \textit{Multi-Oriented (MO)}, \textit{Artistic}, \textit{Contextless}, \textit{Salient}, \textit{Multi-Words} and \textit{General}; 3) occluded scene text dataset (\textit{OST})~\cite{Wang_2021_visionlan}, which is categorized into two subsets based on the degree of occlusion: weak occlusion (\textit{OST}$_w$) and heavy occlusion (\textit{OST}$_h$); 4) long text benchmark (\textit{LTB})~\cite{du2024smtr}, which includes 3376 samples of text length from 25 to 35; 5) the test set of BCTR~\cite{chen2021benchmarking}, a Chinese text recognition benchmark with four subsets: \textit{Scene}, \textit{Web}, \textit{Document} (\textit{Doc}) and \textit{Hand-Writing} (\textit{HW}).

For English recognition, there are three large-scale real-world training sets, i.e., the \textit{Real} dataset~\cite{BautistaA22PARSeq}, \textit{REBU-Syn}~\cite{Rang_2024_CVPR_clip4str}, and \textit{Union14M-L}~\cite{jiang2023revisiting}. However, they all overlap with \textit{U14M} (detailed in \textit{Suppl.~Sec.}~8) across the seven subsets, leading to data leakage, which makes them unsuitable for training models. To resolve this, we introduce a filtered version of \textit{Union14M-L}, termed as \textit{U14M-Filter}, by filtering these overlapping instances. This new dataset is used to train SVTRv2 and 24 popular STR methods.

For Chinese recognition, we train models on the training set of \textit{BCTR}~\cite{chen2021benchmarking}. Unlike previous methods that train separately for each subset, we trained the model on their integration and then evaluated it on the four subsets.


We use AdamW optimizer~\cite{adamw} with a weight decay of 0.05 for training. The LR is set to $6.5\times 10^{-4}$ and batchsize is set to 1024. One cycle LR scheduler~\cite{cosine} with 1.5/4.5 epochs linear warm-up is used in all the 20/100 epochs, where a/b means a for English and b for Chinese. For English models, the training is conducted in two phases: firstly without SGM and then with SGM, both using the above settings. Word accuracy is used as the evaluation metric. Data augmentation like rotation, perspective distortion, motion blur, and gaussian noise, are randomly performed. The maximum text length is set to 25 during training. The size of the character set $N_c$ is set to 94 for English and 6624~\cite{ppocrv3} for Chinese. In the experiments below, SVTRv2 means SVTRv2-B unless specified. All models are trained on 4 RTX 4090 GPUs.

\subsection{Ablation Study}

\begin{figure*}[t]
  \centering
\includegraphics[width=0.98\textwidth]{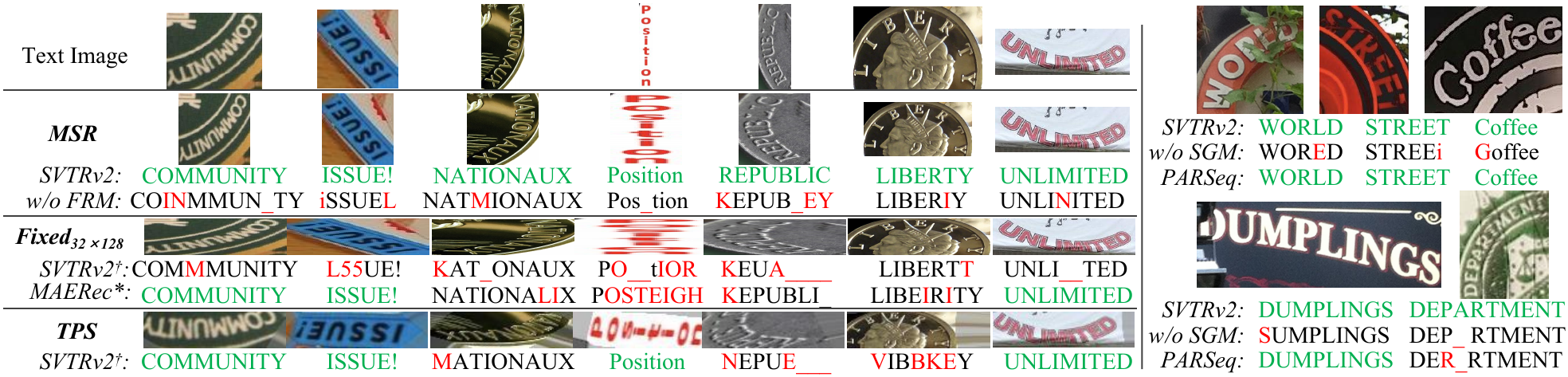} 
    \caption{Qualitative comparison of SVTRv2 with previous methods on irregular and occluded text. $^\dagger$ means that SVTRv2 utilizes the fixed-size (in \textit{Fixed}$_{32\times128}$ part) or rectification module (in \textit{TPS} part) as the resize strategy. \textit{MAERec*} means that SVTRv2$^\dagger$ integrates with the attention-based decoder from the previous best model, i.e. MAERec~\cite{jiang2023revisiting}, such a decoder is widely employed in~\cite{Sheng2019nrtr,pr2021MASTER,cvpr2021TransOCR,xie2022toward_cornertrans,yuICCV2023clipctr,yang2024class_cam,Xu_2024_CVPR_OTE}. \textcolor{green}{Green}, \textcolor{red}{red}, and \textcolor{red}{\_} denotes correctly, wrongly and missed recognition, respectively.}
\label{fig:case}
\end{figure*}

\noindent\textbf{Effectiveness of MSR}. We group \textit{Curve} and \textit{MO} text in \textit{U14M} based on the aspect ratio $R_i$. As shown in Tab.~\ref{tab:msr_FRM}, the majority of irregular texts fall within $R_1$ and $R_2$, where they are particularly prone to distortion when resized to a fixed size (see \textit{Fixed}$_{32\times128}$ in Fig.~\ref{fig:case}). In contrast, MSR demonstrates significant improvements of 15.3\% in $R_1$ and 5.2\% in $R_2$ compared to \textit{Fixed}$_{32\times128}$. Meanwhile, a large fixed-size \textit{Fixed}$_{64\times256}$, although improving the accuracy compared to the baseline, still performs worse than our MSR by clear margins. The results strongly confirm our hypothesis that undesired resizing would hurt the recognition. Our MSR effectively mitigates this issue, providing better visual features thus enhancing the recognition accuracy.

\noindent\textbf{Effectiveness of FRM}. We ablate the two rearrangement sub-modules (Horizontal (H) rearranging and Vertical (V) rearranging). As shown in Tab.~\ref{tab:msr_FRM}, compared to without FRM (w/o FRM), they individually improve accuracy by 2.03\% and 0.71\% on \textit{MO}, and they together result in a 2.46\% gain. In addition, we validate the use of a Transformer Block (+ TF$_1$) as an alternative to splitting the process into two steps for learning the matrix $\mathbf{M}$ holistically. However, its effectiveness is less pronounced, likely because it fails to effectively distinguish between vertical and horizontal orientations. In contrast, FRM performs feature rearrangement in both directions, making it highly sensitive to text irregularity, and thus facilitating accurate CTC alignment. As shown in the left five cases in Fig.~\ref{fig:case}, FRM successfully recognizes reverse instances, providing strong evidence of FRM's effectiveness.

\begin{table}[t]\footnotesize
\centering
\setlength{\tabcolsep}{1pt}{
\begin{tabular}{c|c|cccc|cc|cc}
\toprule

\multicolumn{2}{c|}{}                  & \begin{tabular}[c]{@{}c@{}}$R_1$\\ 2,688\end{tabular} & \begin{tabular}[c]{@{}c@{}}$R_2$\\ 788\end{tabular} & \begin{tabular}[c]{@{}c@{}}$R_3$\\ 266\end{tabular} & \begin{tabular}[c]{@{}c@{}}$R_4$\\ 32\end{tabular} & \textit{Curve} & \textit{MO}    & \textit{Com}   & \textit{U14M}  \\

\midrule

\multicolumn{2}{c|}{SVTRv2 (+MSR+FRM)}                                                              & 87.4 & 88.3 & 86.1 & 87.5 & 88.17 & 86.19 & 96.16 & 83.86 \\
\multicolumn{2}{c|}{SVTRv2 (w/o both)}                                                              & 70.5 & 81.5 & 82.8 & 84.4 & 82.89  & 65.59 & 95.28 & 77.78 \\
\midrule
\multirow{3}{*}{\begin{tabular}[c]{@{}c@{}}vs.\\ MSR \\ (+FRM)\end{tabular}}                                                  & Fixed$_{32\times 128}$         & 72.1 & 83.1 & 84.1 & 85.6 & 83.18 & 68.71 & 95.56 & 78.87 \\
& Padding$_{32\times W}$      & 52.1 & 71.3 & 82.3 & 87.4 & 71.06 & 51.57 & 94.70 & 71.82 \\
& Fixed$_{64\times 256}$         & 76.6 & 81.6 & 81.9 & 80.2 & 85.70 & 67.49 & 95.07 & 79.03 \\
\midrule
\multirow{4}{*}{\begin{tabular}[c]{@{}c@{}}vs.\\ FRM \\ (+MSR)\end{tabular}}                                                 & w/o FRM           & 85.7 & 86.3 & 86.0 & 85.5 & 87.35 & 83.73 & 95.44 & 82.22 \\
& + H rearranging    & 87.0   & 87.1   & 86.3  & 85.5  & 88.05 & 85.76 & 95.98 & 82.94 \\
& + V rearranging   & 85.0   & 87.6   & 88.5  & 85.5  & 88.01 & 84.44 & 95.66 & 82.70 \\     
& + TF$_1$     & 86.4   & 86.3   & 87.5  & 86.1  & 87.51 & 85.50 & 95.60 & 82.49 \\
\bottomrule
\toprule
\multirow{5}{*}{-}                                                   & ResNet+TF$_3$ & 49.3 & 63.5 & 64.0 & 66.7 & 65.00 & 42.07 & 92.26 & 63.00 \\
& FocalNet-B     & 56.7 & 73.2 & 75.3 & 73.9 & 76.46 & 45.80 & 94.49 & 71.63 \\
& ConvNeXtV2   & 58.4 & 71.0 & 73.6 & 71.2 & 75.97 & 45.95 & 93.93 & 70.43 \\
& ViT-S          & 68.5 & 73.8 & 73.8 & 73.0 & 75.02 & 64.35 & 93.57 & 72.09 \\
& SVTR-B         & 53.3 & 74.8 & 76.4 & 78.4 & 76.22 & 44.49 & 94.58 & 71.17 \\
\midrule
\multirow{5}{*}{+FRM}     & ResNet+TF$_3$ & 53.8 & 67.9 & 65.5 & 65.8 & 69.00 & 46.02 & 93.12 & 66.81 \\
& FocalNet-B     & 57.1 & 75.2 & 77.1 & 78.4 & 75.52 & 51.21 & 94.39 & 72.73 \\
& ConvNeXtV2   & 60.7 & 79.0 & 79.0 & 81.1 & 79.72 & 53.32 & 94.19 & 73.09 \\
& ViT-S          & 75.1 & 79.4 & 79.0 & 78.4 & 80.42 & 72.17 & 94.44 & 77.07 \\
& SVTR-B         & 59.1 & 79.0 & 78.8 & 80.2 & 79.84 & 51.28 & 94.75 & 73.48 \\
\midrule
\multirow{3}{*}{+MSR} & ResNet+TF$_3$ & 68.2 & 71.3 & 75.3 & 72.1 & 75.64 & 60.33 & 93.50 & 71.95 \\
& FocalNet-B     & 80.5 & 80.6 & 79.2 & 85.0 & 82.26 & 74.82 & 94.92 & 78.94 \\
& ConvNeXtV2   & 76.2 & 79.0 & 82.3 & 80.2 & 81.05 & 73.27 & 94.60 & 77.71 \\
\midrule
\multicolumn{10}{c}{\setlength{\tabcolsep}{1pt}{
\begin{tabular}{c|ccc|ccc|cc}
 - / + SGM & \textit{OST}$_w$ & \textit{OST}$_h$ & Avg   & \textit{OST}$_w^*$ & \textit{OST}$_h^*$ & Avg   & \textit{Com}$^*$   & \textit{U14M}$^*$  \\
\midrule
ResNet+TF$_3$ & 71.6 & 51.8 & 61.72 & 77.9 & 55.0 & 66.43 & 95.19 & 78.61 \\
FocalNet-B     & 78.9 & 62.8 & 70.88 & 84.6 & 70.6 & 77.61 & 96.28 & 84.10 \\
ConvNeXtV2   & 76.0 & 58.2 & 67.10 & 82.0 & 63.9 & 72.97 & 96.09 & 82.10 \\
\end{tabular}}} \\
\bottomrule
\end{tabular}}
\caption{Ablations on MSR and FRM (top) and assessing MSR, FRM, and SGM across visual models (lower). * means with SGM.}
\label{tab:msr_FRM}
\end{table}

\begin{table}[t]\footnotesize
\centering
\setlength{\tabcolsep}{3pt}{
\begin{tabular}{c|c|ccc|cc}
\toprule
& Method & \textit{OST}$_w$ & \textit{OST}$_h$ & Avg & \textit{Com} & \textit{U14M} \\
\midrule
\multirow{7}{*}{\begin{tabular}[c]{@{}c@{}}Linguistic \\context \\ modeling\end{tabular}} & w/o SGM      & 82.86          & 66.97          & 74.92     & 96.16 & 83.86     \\
& SGM      & \textbf{86.26} & \textbf{73.80}          & \textbf{80.03}  & \textbf{96.57} & \textbf{86.14} \\
& GTC~\cite{hu2020gtc}        &      83.07          &     68.32           &  75.70  & 96.01 & 84.33            \\
& ABINet~\cite{fang2021abinet}      &      83.07          &     67.54           &      75.31  & 96.25 & 84.17        \\
& VisionLAN~\cite{Wang_2021_visionlan}     &   83.25             &  68.97              &      76.11 & 96.39 & 84.01         \\
& PARSeq~\cite{BautistaA22PARSeq} &      83.85          &     69.24    &   76.55    & 96.21 & 84.72 \\
& MAERec~\cite{jiang2023revisiting}       &       83.21         &     69.69           &  76.45   & 96.47 & 84.69           \\
\bottomrule
\end{tabular}}
\caption{Comparison of the proposed SGM with other language models in linguistic context modeling on \textit{OST}. }
\label{tab:semantic}
\end{table}

\begin{table*}[t]\footnotesize
\centering
\setlength{\tabcolsep}{1.8pt}{
\begin{tabular}{cc|c|c|ccccccc|cccccccc|c|c|c|c}
\multicolumn{23}{c}{\setlength{\tabcolsep}{3.5pt}{\begin{tabular}{
>{\columncolor[HTML]{FFFFC7}}c 
>{\columncolor[HTML]{FFFFC7}}c 
>{\columncolor[HTML]{FFFFC7}}c 
>{\columncolor[HTML]{FFFFC7}}c 
>{\columncolor[HTML]{FFFFC7}}c 
>{\columncolor[HTML]{FFFFC7}}c c
>{\columncolor[HTML]{ECF4FF}}c 
>{\columncolor[HTML]{ECF4FF}}c 
>{\columncolor[HTML]{ECF4FF}}c 
>{\columncolor[HTML]{ECF4FF}}c 
>{\columncolor[HTML]{ECF4FF}}c 
>{\columncolor[HTML]{ECF4FF}}c 
>{\columncolor[HTML]{ECF4FF}}c }
\toprule
\textit{IIIT5k} & \textit{SVT} & \textit{ICDAR2013} & \textit{ICDAR2015} & \textit{SVTP} & \textit{CUTE80} & $\|$ & \textit{Curve} & \textit{Multi-Oriented} & \textit{Artistic} & \textit{Contextless} & \textit{Salient} & \textit{Multi-Words} & \textit{General} 
\end{tabular}}} \\
\toprule
\multicolumn{2}{c|}{Method}                         & Venue              & Encoder        & \multicolumn{6}{c}{\cellcolor[HTML]{FFFFC7}Common Benchmarks (\textit{Com})} & Avg   & \multicolumn{7}{c}{\cellcolor[HTML]{ECF4FF}Union14M-Benchmark (\textit{U14M})} & Avg   & \textit{LTB} & \textit{OST} & \textit{Size} & \textit{FPS}  \\
\midrule
\multicolumn{2}{r|}{ASTER~\cite{shi2019aster}}                          & TPAMI19          & ResNet+LSTM  & 96.1     & 93.0     & 94.9      & 86.1     & 87.9     & 92.0    & 91.70  & 70.9    & 82.2    & 56.7    & 62.9    & 73.9    & 58.5   & 76.3   & 68.75 & 0.1 & 61.9 &19.0 & 67.1 \\
\multicolumn{2}{r|}{NRTR~\cite{Sheng2019nrtr}}                           & ICDAR19          & Stem+TF$_6$   & 98.1     & 96.8     & 97.8     & 88.9     & 93.3     & 94.4    & 94.89 & 67.9    & 42.4    & 66.5    & 73.6    & 66.4    & 77.2   & 78.3   & 67.46 & 0.0 & 74.8 &44.3 & 17.3 \\
\multicolumn{2}{r|}{MORAN~\cite{pr2019MORAN}}                          & PR19             & ResNet+LSTM  & 96.7     & 91.7     & 94.6     & 84.6     & 85.7     & 90.3    & 90.61 & 51.2    & 15.5    & 51.3    & 61.2    & 43.2    & 64.1   & 69.3   & 50.82 & 0.1 & 57.9 &17.4 & 59.5 \\
\multicolumn{2}{r|}{SAR~\cite{li2019sar}}                            & AAAI19           & ResNet+LSTM  & 98.1     & 93.8     & 96.7     & 86.0     & 87.9     & 95.5    & 93.01 & 70.5    & 51.8    & 63.7    & 73.9    & 64.0    & 79.1   & 75.5   & 68.36 & 0.0 & 60.6 &57.5 & 15.8 \\
\multicolumn{2}{r|}{DAN~\cite{wang2020aaai_dan}}                            & AAAI20           & ResNet+FPN   & 97.5     & 94.7     & 96.5     & 87.1     & 89.1     & 94.4    & 93.24 & 74.9    & 63.3    & 63.4    & 70.6    & 70.2    & 71.1   & 76.8   & 70.05 & 0.0 & 61.8 &27.7 & 99.0 \\
\multicolumn{2}{r|}{SRN~\cite{yu2020srn}}                            & CVPR20           & ResNet+FPN   & 97.2    & 96.3      & 97.5     & 87.9     & 90.9     & 96.9    & 94.45 & 78.1    & 63.2    & 66.3    & 65.3    & 71.4    & 58.3   & 76.5   & 68.43 & 0.0 & 64.6 &51.7 & 67.1 \\
\multicolumn{2}{r|}{SEED~\cite{cvpr2020seed}}                            & CVPR20           & ResNet+LSTM   & 96.5     & 93.2    & 94.2    & 87.5    & 88.7   & 93.4        & 92.24    & 69.1   & 80.9   & 56.9     & 63.9     & 73.4    & 61.3   & 76.5   & 68.87 & 0.1 & 62.6 &24.0 & 65.4\\
\multicolumn{2}{r|}{AutoSTR~\cite{zhang2020autostr}}                        & ECCV20           & NAS+LSTM     & 96.8     & 92.4     & 95.7     & 86.6     & 88.2     & 93.4    & 92.19 & 72.1    & 81.7    & 56.7    & 64.8    & 75.4    & 64.0   & 75.9   & 70.09 & 0.1 & 61.5 &6.0  & 82.6\\
\multicolumn{2}{r|}{RoScanner~\cite{yue2020robustscanner}}                      & ECCV20           & ResNet       & 98.5     & 95.8     & 97.7     & 88.2     & 90.1     & 97.6    & 94.65 & 79.4    & 68.1    & 70.5    & 79.6    & 71.6    & 82.5   & 80.8   & 76.08 & 0.0 & 68.6 &48.0 & 64.1 \\
\multicolumn{2}{r|}{ABINet~\cite{fang2021abinet}}                         & CVPR21           & ResNet+TF$_3$ & 98.5     & 98.1     & 97.7     & 90.1     & 94.1     & 96.5    & 95.83 & 80.4    & 69.0    & 71.7    & 74.7    & 77.6    & 76.8   & 79.8   & 75.72 & 0.0 & 75.0 &36.9 & 73.0 \\
\multicolumn{2}{r|}{VisionLAN~\cite{Wang_2021_visionlan}}                      & ICCV21           & ResNet+TF$_3$ & 98.2     & 95.8     & 97.1     & 88.6     & 91.2     & 96.2    & 94.50 & 79.6    & 71.4    & 67.9    & 73.7    & 76.1    & 73.9   & 79.1   & 74.53 & 0.0 & 66.4 &32.9 & 93.5 \\
\multicolumn{2}{r|}{PARSeq~\cite{BautistaA22PARSeq}}                        & ECCV22           & ViT-S          & 98.9     & 98.1     & 98.4     & 90.1     & 94.3     & 98.6    & 96.40 & 87.6    & 88.8    & 76.5    & 83.4    & 84.4    & 84.3   & 84.9   & 84.26 & 0.0 & 79.9 &23.8 & 52.6 \\
\multicolumn{2}{r|}{MATRN~\cite{MATRN}}                          & ECCV22           & ResNet+TF$_3$ & 98.8     & 98.3     & 97.9     & 90.3     & 95.2     & 97.2    & 96.29 & 82.2    & 73.0    & 73.4    & 76.9    & 79.4    & 77.4   & 81.0   & 77.62 & 0.0 & 77.8 &44.3 & 46.9\\
\multicolumn{2}{r|}{MGP-STR~\cite{mgpstr}}                        & ECCV22           & ViT-B          & 97.9 &	97.8 &	97.1 	&89.6 	&95.2 	&96.9 	&95.75 	&85.2 	&83.7 	&72.6 	&75.1 	&79.8 	&71.1 	&83.1 	&78.65  & 0.0 & 78.7 & 148 & 120\\

\multicolumn{2}{r|}{LPV~\cite{ijcai2023LPV}}                          & IJCAI23          & SVTR-B         & 98.6     & 97.8     & 98.1     & 89.8     & 93.6     & 97.6    & 95.93 & 86.2    & 78.7    & 75.8    & 80.2    & 82.9    & 81.6   & 82.9   & 81.20 & 0.0 & 77.7 & 30.5 & 82.6\\

\multicolumn{2}{r|}{MAERec~\cite{jiang2023revisiting}}                         & ICCV23           & ViT-S & \textbf{99.2}               & 97.8                        & 98.2                        & 90.4                        & 94.3                        & 98.3                        & 96.36                        & 89.1                        & 87.1                        & 79.0                        & 84.2                        & \textbf{86.3}                                 & 85.9                        & 84.6                        & 85.17                   & 9.8 & 76.4     & 35.7    & 17.1                                            \\

\multicolumn{2}{r|}{LISTER~\cite{iccv2023lister}}                        & ICCV23           & FocalNet-B     & 98.8     & 97.5     & 98.6     & 90.0     & 94.4     & 96.9    & 96.03 & 78.7    & 68.8    & 73.7    & 81.6    & 74.8    & 82.4   & 83.5   & 77.64 & 36.3 & 77.1 &51.1 & 44.6 \\
\multicolumn{2}{r|}{CDistNet~\cite{zheng2024cdistnet}}                       & IJCV24           & ResNet+TF$_3$ & 98.7     & 97.1     & 97.8     & 89.6     & 93.5     & 96.9    & 95.59 & 81.7    & 77.1    & 72.6    & 78.2    & 79.9    & 79.7   & 81.1   & 78.62 & 0.0 & 71.8 &43.3 & 15.9\\
\multicolumn{2}{r|}{CAM~\cite{yang2024class_cam}}                            & PR24             & ConvNeXtV2   & 98.2     & 96.1     & 96.6     & 89.0     & 93.5     & 96.2    & 94.94 & 85.4    & 89.0    & 72.0    & 75.4    & 84.0    & 74.8   & 83.1   & 80.52 & 0.7 & 74.2 & 58.7 & 28.6\\
\multicolumn{2}{r|}{BUSNet~\cite{Wei_2024_busnet}}                         & AAAI24           & ViT-S          & 98.3     & 98.1     & 97.8     & 90.2     & \textbf{95.3}     & 96.5    & 96.06 & 83.0    & 82.3    & 70.8    & 77.9    & 78.8    & 71.2   & 82.6   & 78.10 & 0.0 & 78.7 &32.1 & 83.3\\
\multicolumn{2}{r|}{OTE~\cite{Xu_2024_CVPR_OTE}}                            & CVPR24           & SVTR-B         & 98.6     & 96.6     & 98.0     & 90.1     & 94.0     & 97.2    & 95.74 & 86.0    & 75.8    & 74.6    & 74.7    & 81.0    & 65.3   & 82.3   & 77.09 & 0.0 & 77.8 &20.3 & 55.2\\
\multicolumn{2}{r|}{CPPD~\cite{du2023cppd}}                         & TPAMI25           & SVTR-B         & 99.0     & 97.8     & 98.2     & 90.4     & 94.0     & \textbf{99.0}    & 96.40 & 86.2    & 78.7    & 76.5    & 82.9    & 83.5    & 81.9   & 83.5   & 81.91 & 0.0 & 79.6 &27.0 & 125 \\
\multicolumn{2}{r|}{IGTR-AR~\cite{du2024igtr}}  & TPAMI25 & SVTR-B        &   98.7    &  \textbf{98.4}    &   98.1    &   90.5     & 94.9     &    98.3  & 96.48                                                                       &  90.4     &     \textbf{91.2}      & 77.0      &   82.4          &     84.7    &  84.0                                                      &  84.4       & 84.86 & 0.0 & 76.3 &    24.1   & 58.3 \\
\multicolumn{2}{r|}{SMTR~\cite{du2024smtr}}  & AAAI25 & FocalSVTR     & 99.0  & 97.4 & 98.3 & 90.1 & 92.7          & 97.9 & 95.90          & 89.1 & 87.7 & 76.8 & 83.9 & 84.6 & \textbf{89.3}          & 83.7 & 85.00       &  \textbf{55.5} & 73.5    & 15.8 &   66.2   \\
\midrule
\multicolumn{1}{c|}{}& CRNN~\cite{shi2017crnn}                     & TPAMI16          & ResNet+LSTM  & 95.8     & 91.8     & 94.6     & 84.9     & 83.1     & 91.0    & 90.21 & 48.1    & 13.0    & 51.2    & 62.3    & 41.4    & 60.4   & 68.2   & 49.24 & 47.2 &58.0 & 16.2 & 172\\
\multicolumn{1}{c|}{}&  SVTR~\cite{duijcai2022svtr}   & IJCAI22          & SVTR-B         & 98.0     & 97.1     & 97.3     & 88.6     & 90.7     & 95.8    & 94.58 & 76.2    & 44.5    & 67.8    & 78.7    & 75.2    & 77.9   & 77.8   & 71.17 & 45.1 &69.6 & 18.1 & 161\\
\multicolumn{1}{c|}{}  & \cellcolor[HTML]{EFEFEF}                         & \multicolumn{1}{c|}{\cellcolor[HTML]{EFEFEF}}                   & \cellcolor[HTML]{EFEFEF}SVTRv2-T  & \cellcolor[HTML]{EFEFEF}98.6          & \cellcolor[HTML]{EFEFEF}96.6          & \cellcolor[HTML]{EFEFEF}98.0          & \cellcolor[HTML]{EFEFEF}88.4          & \cellcolor[HTML]{EFEFEF}90.5 & \cellcolor[HTML]{EFEFEF}96.5          & \cellcolor[HTML]{EFEFEF}94.78          & \cellcolor[HTML]{EFEFEF}83.6          & \cellcolor[HTML]{EFEFEF}76.0          & \cellcolor[HTML]{EFEFEF}71.2          & \cellcolor[HTML]{EFEFEF}82.4          & \cellcolor[HTML]{EFEFEF}77.2 & \cellcolor[HTML]{EFEFEF}82.3          & \cellcolor[HTML]{EFEFEF}80.7          & \cellcolor[HTML]{EFEFEF}79.05 & \cellcolor[HTML]{EFEFEF}47.8         & \cellcolor[HTML]{EFEFEF}71.4 & \cellcolor[HTML]{EFEFEF}5.1  & \cellcolor[HTML]{EFEFEF}201\\
\multicolumn{1}{c|}{}     & \cellcolor[HTML]{EFEFEF}                         & \multicolumn{1}{c|}{\cellcolor[HTML]{EFEFEF}}                   & \cellcolor[HTML]{EFEFEF}SVTRv2-S & \cellcolor[HTML]{EFEFEF}99.0          & \cellcolor[HTML]{EFEFEF}98.3 & \cellcolor[HTML]{EFEFEF}98.5          & \cellcolor[HTML]{EFEFEF}89.5          & \cellcolor[HTML]{EFEFEF}92.9 & \cellcolor[HTML]{EFEFEF}98.6          & \cellcolor[HTML]{EFEFEF}96.13          & \cellcolor[HTML]{EFEFEF}88.3          & \cellcolor[HTML]{EFEFEF}84.6          & \cellcolor[HTML]{EFEFEF}76.5          & \cellcolor[HTML]{EFEFEF}84.3          & \cellcolor[HTML]{EFEFEF}83.3 & \cellcolor[HTML]{EFEFEF}85.4          & \cellcolor[HTML]{EFEFEF}83.5          & \cellcolor[HTML]{EFEFEF}83.70 & \cellcolor[HTML]{EFEFEF}47.6        & \cellcolor[HTML]{EFEFEF}78.0 & \cellcolor[HTML]{EFEFEF}11.3 & \cellcolor[HTML]{EFEFEF}189\\

\multicolumn{1}{c|}{\multirow{-5}{*}{\begin{tabular}[c]{@{}c@{}}C\\ T\\ C\end{tabular}}} & \multirow{-3}{*}{\cellcolor[HTML]{EFEFEF}SVTRv2} & \multicolumn{1}{c|}{\multirow{-3}{*}{\cellcolor[HTML]{EFEFEF}-}} & \cellcolor[HTML]{EFEFEF}SVTRv2-B  & \cellcolor[HTML]{EFEFEF}\textbf{99.2} & \cellcolor[HTML]{EFEFEF}98.0 & \cellcolor[HTML]{EFEFEF}\textbf{98.7} & \cellcolor[HTML]{EFEFEF}\textbf{91.1} & \cellcolor[HTML]{EFEFEF}93.5 & \cellcolor[HTML]{EFEFEF}\textbf{99.0} & \cellcolor[HTML]{EFEFEF}\textbf{96.57} & \cellcolor[HTML]{EFEFEF}\textbf{90.6} & \cellcolor[HTML]{EFEFEF}89.0 & \cellcolor[HTML]{EFEFEF}\textbf{79.3} & \cellcolor[HTML]{EFEFEF}\textbf{86.1} & \cellcolor[HTML]{EFEFEF}86.2 & \cellcolor[HTML]{EFEFEF}86.7 & \cellcolor[HTML]{EFEFEF}\textbf{85.1} & \cellcolor[HTML]{EFEFEF}\textbf{86.14} & \cellcolor[HTML]{EFEFEF}50.2 & \cellcolor[HTML]{EFEFEF}\textbf{80.0} & \cellcolor[HTML]{EFEFEF}19.8 & \cellcolor[HTML]{EFEFEF}143
    \\
\bottomrule
\end{tabular}}
\caption{All the models and SVTRv2 are trained on \textit{U14M-Filter}. To ensuring that the results reflect the true potential of these methods under their best experimental settings, we conducted extensive tuning (detailed in \textit{Suppl.~Sec.}~12) of the model-specific settings (e.g., optimizer, learning rate, and regularization) and reported the best result we got. TF$_n$ denotes the $n$-layer Transformer block~\cite{NIPS2017_attn}. \textit{Size} denotes the number of parameters of the model ($\times 10^6$). \textit{FPS} is measured on one NVIDIA 1080Ti GPU.}
\label{tab:sota}
\end{table*}

\noindent\textbf{Effectiveness of SGM}. As illustrated in Tab.~\ref{tab:semantic}, SGM achieves 0.41\% and 2.28\% increase on \textit{Com} and \textit{U14M}, respectively, while gains a 5.11\% improvement on \textit{OST}. Since \textit{OST} frequently suffers from missing a portion of characters, this notable gain implies that the linguistic context has been successfully established. For comparison, we also employ GTC~\cite{hu2020gtc} and four popular language decoders~\cite{fang2021abinet,Wang_2021_visionlan,BautistaA22PARSeq,jiang2023revisiting} to substitute for our SGM. However, there is no much difference between the gains obtained from \textit{OST} and the other two datasets (\textit{Com} and \textit{U14M}). This suggests that SGM offers a distinct advantage in integrating linguistic context into visual features, and significantly improving the recognition accuracy of CTC models. The five cases on the right side of Fig.~\ref{fig:case} showcase that SGM facilitates SVTRv2 to accurately decipher occluded characters, achieving comparable results with PARSeq~\cite{BautistaA22PARSeq}, which is equipped with an advanced permuted language model.

\noindent\textbf{Adaptability to different visual models.} We further examine MSR, FRM, and SGM on five frequently used visual models~\cite{he2016resnet,dosovitskiy2020vit,duijcai2022svtr,YangLDG22focalnet,WooDHC0KX23_ConvNeXtv2}. As presented in the bottom part of Tab.~\ref{tab:msr_FRM}, these modules consistently enhance the performance (ViT~\cite{dosovitskiy2020vit} and SVTR~\cite{duijcai2022svtr} employ absolute positional coding and do not compatible with MSR). When both FRM and MSR modules incorporated, ResNet+TF$_3$~\cite{he2016resnet}, FocalNet~\cite{YangLDG22focalnet}, and ConvNeXtV2~\cite{yang2024class_cam} exhibit significant accuracy improvements, either matching or even exceeding the accuracy of their EDTR counterparts (see Tab.~\ref{tab:sota}). The results highlight the versatility of the three proposed modules.

\subsection{Comparison with State-of-the-arts}
We compare SVTRv2 with 24 popular STR methods on \textit{Com}, \textit{U14M}, \textit{OST}, and \textit{LTB}. The results are presented in Tab.~\ref{tab:sota}. SVTRv2-B achieves top results in 9 out of the 15 evaluated scenarios and outperforms the most of EDTRs, showing a clear accuracy advantage. Meanwhile, it enjoys a small model size and a significant speed advantage. Specifically, compared to MAERec, the best-performed existing model on \textit{U14M}, SVTRv2-B shows an accuracy improvement of 0.97\% and 8$\times$ faster inference speed. Compared to CPPD, which is known for its wonderful accuracy-speed tradeoff, SVTRv2-B runs faster than 10\%, along with a 4.23\% accuracy increase on \textit{U14M}. Regarding \textit{OST}, as illustrated in the right part of Fig.~\ref{fig:case}, SVTRv2-B relies solely on a single visual model but achieves comparable accuracy to PARSeq, which employed the advanced permuted language model and is the best-performed existing model on \textit{OST}. In addition, SVTRv2-T and SVTRv2-S, the two smaller models also show leading accuracy compared with models of similar sizes, offering flexible solutions with different accuracy-speed tradeoff.

\begin{figure}[t]
  \centering
\includegraphics[width=0.47\textwidth]{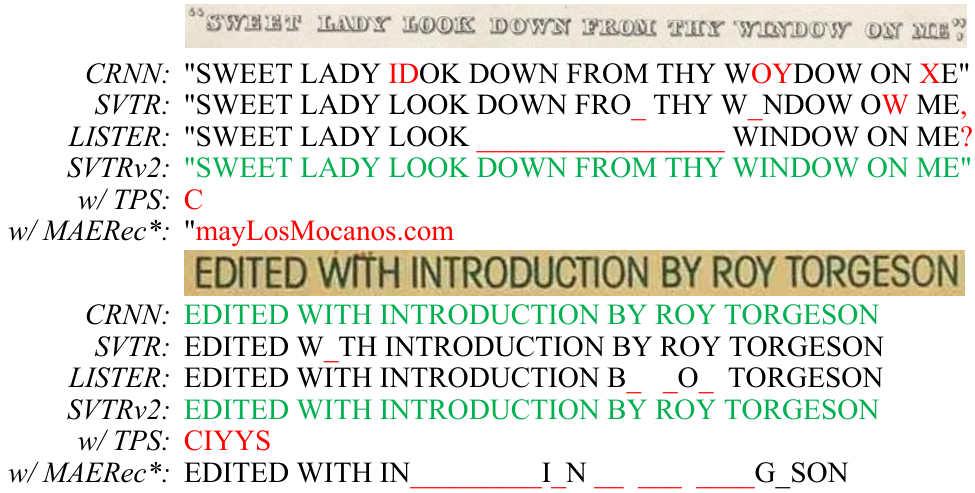} 
\caption{Long text recognition examples. \textit{TPS} and \textit{MAERec*} denote SVTRv2 integrated with TPS and the decoder of MAERec.}
\label{fig:case_long}
\end{figure}

Two observations are derived by looking at the results on \textit{Curve} and \textit{MO}. First, SVTRv2 models significantly surpass existing CTC models. For example, compared to SVTR-B, SVTRv2-B gains prominent accuracy improvements of 14.4\% and 44.5\% on the two subsets, respectively. Second, as shown in Tab.~\ref{tab:tps_decoder},
comparing to previous methods employing rectification modules~\cite{shi2019aster,cvpr2020seed,zhang2020autostr,zheng2024cdistnet,yang2024class_cam,duijcai2022svtr,zheng2023tps++} or attention-based decoders ~\cite{Sheng2019nrtr,cvpr2021TransOCR,xie2022toward_cornertrans,yuICCV2023clipctr,jiang2023revisiting,yang2024class_cam,Xu_2024_CVPR_OTE,wang2020aaai_dan,li2019sar} to recognize irregular text, SVTRv2 also performs better than these methods on \textit{Curve}. In Fig.~\ref{fig:case}, \textit{TPS} (a rectification module) and \textit{MAERec*} (an attention-based decoder) do not recognize the extremely curved and rotated text correctly. In contrast, SVTRv2 successes. Moreover, as demonstrated by the results on \textit{LTB} in Tab.~\ref{tab:tps_decoder} and Fig.~\ref{fig:case_long}, \textit{TPS} and \textit{MAERec*} both do not effectively recognize long text, while SVTRv2 circumvents this limitation. These results indicate that our proposed modules successfully address the challenge of handling irregular text that existing CTC models encountered, while still preserving CTC's proficiency in recognizing long text.

SVTRv2 also exhibit strong performance in Chinese text recognition (see Tab.~\ref{tab:ch_all}), where SVTRv2-B achieve state of the art. The result implies its great adaptability to different languages. Moreover, it also shows superior performance on Chinese long text (\textit{Scene}$_{L>25}$). To sum, we evaluate SVTRv2 across a wide range of scenarios. The results consistently confirm that this CTC model beats leading EDTRs.

\begin{table}[t]\footnotesize
\centering
\setlength{\tabcolsep}{1.5pt}{
\begin{tabular}{c|c|cccc|cc|cc|c}
\toprule
\multicolumn{2}{c|}{}                 & $R_1$ & $R_2$ & $R_3$ & $R_4$ & \textit{Curve} & \textit{MO}    & \textit{Com}   & \textit{U14M}  & \textit{LTB}  \\
\midrule
\multicolumn{2}{c|}{SVTRv2}                                                     & \textbf{90.8} & \textbf{89.0} & \textbf{90.4} & \textbf{91.0} & \textbf{90.64} & \textbf{89.04} & \textbf{96.57} & \textbf{86.14} & \textbf{50.2} \\
\midrule    
\multirow{2}{*}{TPS} & SVTR~\cite{duijcai2022svtr} & 86.8 & 82.3 & 77.3 & 75.7 & 82.19 & 86.12 & 94.62 & 78.44 & 0.0\\
 & SVTRv2 & 89.5 & 85.1 & 78.4 & 83.8 & 84.71 & 88.97 & 94.62 & 79.94 & 0.5\\
\midrule
\multirow{2}{*}{\begin{tabular}[c]{@{}c@{}}MAE-\\REC*\end{tabular}} & SVTR~\cite{duijcai2022svtr} & 81.3 & 87.6 & 87.6 & 88.3 & 87.88 & 78.74 & 96.32 & 83.23 & 0.0\\
 & SVTRv2 & 88.0 & 88.9 & 89.4 & 88.3 & 89.96 & 87.56 & 96.42 & 85.67  & 0.2\\
\bottomrule
\end{tabular}}
\caption{SVTRv2 and SVTR comparisons on irregular text and \textit{LTB}, where the rectification module (TPS) and the attention-based decoder (MAERec*) are employed.}
\label{tab:tps_decoder}
\end{table}

\begin{table}[t]\footnotesize
\centering
\setlength{\tabcolsep}{3pt}{
\begin{tabular}{r|ccccc|c|c}
\toprule
Method        & \textit{Scene}         & \textit{Web}           & \textit{Doc} & \textit{HW} & Avg         & \textit{Scene}$_{L>25}$   & \textit{Size} \\
\midrule
ASTER~\cite{shi2019aster}        & 61.3          & 51.7          & 96.2          & 37.0            & 61.55    &  -    & 27.2   \\
MORAN~\cite{pr2019MORAN}        & 54.6          & 31.5          & 86.1          & 16.2          & 47.10    &   -    & 28.5   \\
SAR~\cite{li2019sar}          & 59.7          & 58.0            & 95.7          & 36.5          & 62.48    &   -   & 27.8   \\
SEED~\cite{cvpr2020seed}         & 44.7          & 28.1          & 91.4          & 21.0            & 46.30     &  -    & 36.1   \\
MASTER~\cite{pr2021MASTER}       & 62.8          & 52.1          & 84.4          & 26.9          & 56.55     &   -  & 62.8   \\
ABINet~\cite{fang2021abinet}       & 66.6          & 63.2          & 98.2          & 53.1          & 70.28     &  -   & 53.1   \\
TransOCR~\cite{cvpr2021TransOCR}     & 71.3          & 64.8          & 97.1          & 53.0            & 71.55    &   -   & 83.9   \\
CCR-CLIP~\cite{yuICCV2023clipctr}     & 71.3          & 69.2          & 98.3          & 60.3          & 74.78      &  -  & 62.0     \\
DCTC~\cite{Zhang_Lu_Liao_Huang_Li_Wang_Peng_2024_DCTC}     & 73.9          & 68.5          & 99.4          & 51.0          & 73.20     &  -   & 40.8     \\
CAM~\cite{yang2024class_cam}     & 76.0          & 69.3          & 98.1          & 59.2          & 76.80     &  -   & 135     \\
PARSeq*~\cite{BautistaA22PARSeq} & 84.2 & 82.8 & \textbf{99.5} & 63.0 & 82.37 &0.0 &28.9
\\

MAERec*~\cite{jiang2023revisiting} & \textbf{84.4} & 83.0 & \textbf{99.5} & 65.6 & 83.13 & 4.1 & 40.8\\
LISTER*~\cite{iccv2023lister} & 79.4 & 79.5 & 99.2 & 58.0 & 79.02 & 13.9 & 55.0 \\
DPTR*~\cite{zhao_2024_acmmm_dptr} & 80.0 & 79.6 & 98.9 & 64.4  & 80.73 & 0.0 &68.0 \\
CPPD*~\cite{du2023cppd} & 82.7 	& 82.4 	& 99.4 	& 62.3 	&81.72 & 0.0& 32.1\\
IGTR-AR*~\cite{du2024igtr}     & 82.0 & 81.7 & \textbf{99.5} &  63.8 &  81.74  & 0.0 & 29.2 \\
SMTR*~\cite{du2024smtr}       & 83.4          &   83.0            &  99.3             &    65.1           &  82.68       &  49.4     &   20.8    \\
\midrule
CRNN*~\cite{shi2017crnn} & 63.8 & 68.2 & 97.0 & 
 46.1 & 68.76 & 37.6 & 19.5 \\ 

SVTR-B*~\cite{duijcai2022svtr} & 77.9 	& 78.7 &	99.2 	&62.1 	&79.49 & 22.9 &19.8 \\
\cellcolor[HTML]{EFEFEF}SVTRv2-T & \cellcolor[HTML]{EFEFEF}77.8 &	\cellcolor[HTML]{EFEFEF}78.8 &	\cellcolor[HTML]{EFEFEF}99.2 &	\cellcolor[HTML]{EFEFEF}62.0 &	\cellcolor[HTML]{EFEFEF}79.45 &   \cellcolor[HTML]{EFEFEF}47.8 &   \cellcolor[HTML]{EFEFEF}6.8\\
\cellcolor[HTML]{EFEFEF}SVTRv2-S & \cellcolor[HTML]{EFEFEF}81.1 &	\cellcolor[HTML]{EFEFEF}81.2 &	\cellcolor[HTML]{EFEFEF}99.3 &	\cellcolor[HTML]{EFEFEF}65.0& 	\cellcolor[HTML]{EFEFEF}81.64 &   \cellcolor[HTML]{EFEFEF}50.0 &   \cellcolor[HTML]{EFEFEF}14.0\\
\cellcolor[HTML]{EFEFEF}SVTRv2-B  &  \cellcolor[HTML]{EFEFEF}83.5 &	\cellcolor[HTML]{EFEFEF}\textbf{83.3} &	\cellcolor[HTML]{EFEFEF}\textbf{99.5} &	\cellcolor[HTML]{EFEFEF}\textbf{67.0} &	\cellcolor[HTML]{EFEFEF}\textbf{83.31}             &   \cellcolor[HTML]{EFEFEF}\textbf{52.8} &   \cellcolor[HTML]{EFEFEF}22.5    \\
\bottomrule
\end{tabular}}
\caption{Results on Chinese text dataset. * denotes that the model is retrained using the same setting as SVTRv2 (\textit{Sec.}~4.1).}
\label{tab:ch_all}
\end{table}

\begin{table}[t]\footnotesize
\centering
\setlength{\tabcolsep}{1.3pt}{
\begin{tabular}{r|cccccc|c|c|c|c}
\toprule
 Method      &  \multicolumn{6}{c|}{Common Benchmarks (\textit{Com})}    & Avg & \textit{OST}  & \textit{Size} & \textit{FPS} \\
\midrule
E$^2$STR~\cite{Zhao_2024_CVPR_E2STR}   & 99.2 & 98.6 & 98.7 & \textbf{93.8} & 96.7 & 99.3 & 97.71 & 80.7 & 211 & 7.86\\
 VL-Reader~\cite{zhong_2024_acmmm_vlreader}  & \textbf{99.6} & 99.1 & 98.7 & 92.6 & 97.5 & 99.3 & 97.80 & 86.2 & 142 & -\\
 CLIP4STR~\cite{zhao_2025_tip_clip4str}    & 99.4 & 98.6 & 98.3 & 90.8 & \textbf{97.8} & 99.0   & 97.32 & 82.8 & 158 & 14.1\\
 DPTR~\cite{zhao_2024_acmmm_dptr}    & 99.5 & \textbf{99.2} & 98.5 & 91.8 & 97.1 & 98.6   & 97.45 & - & 66.5 & 49.3 \\
 IGTR~\cite{du2024igtr}   & 99.2      &   98.3   &   \textbf{98.8}   &  92.0    &  96.8    &   99.0   & 97.34 & 86.5  & 24.1 & 58.3 \\
 \midrule
 SVTRv2-B             & 99.2 &	98.6 &	\textbf{98.8} &	\textbf{93.8} &	97.2 &	\textbf{99.4} &  \textbf{97.83} &	\textbf{86.9} & 19.8 & 143\\

\bottomrule
\end{tabular}}
\caption{Quantitative comparison of SVTRv2 with four advanced EDTRs experienced large-scale vision-language pretraining. For fairness, SVTRv2 is fine-tuned on \textit{Real} dataset to align with them.}
\label{tab:add_result}
\end{table}

In addition, recent EDTRs advances, e.g., E$^2$STR~\cite{Zhao_2024_CVPR_E2STR}, VL-Reader~\cite{zhong_2024_acmmm_vlreader}, CLIP4STR~\cite{zhao_2025_tip_clip4str}, and DPTR~\cite{zhao_2024_acmmm_dptr}, achieve impressive accuracy through large-scale vision-language pretraining. To align with these methods, we conduct an experiment by adding pretraining to SVTRv2 on synthetic datasets \cite{Synthetic,jaderberg14synthetic} and fine-tuning on \textit{Real} dataset \cite{BautistaA22PARSeq}. The results in Tab.~\ref{tab:add_result} show that this pretraining significantly enhances SVTRv2's performance, allowing it to surpass the aforementioned models. Notably, SVTRv2 achieves the highest average accuracy in \textit{Com} (97.8\%) while also demonstrating superior generalization to \textit{OST} (86.9\%). Compared to CLIP4STR, SVTRv2 achieves these results with only 14\% of the parameters and runs 10$\times$ faster, highlighting its efficiency. These findings again validate the effectiveness of our SVTRv2, as well as the proposed strategies or modules, i.e., MSR, FRM, and SGM.

\section{Conclusion}

In this paper, we have presented SVTRv2, an accurate and efficient CTC-based STR method. SVTRv2 is featured by developing the MSR and FRM modules to tackle the text irregular challenge, and devising the SGM module to endow linguistic context to the visual model. These upgrades maintain the simple inference architecture of CTC models, thus they remain quite efficient. More importantly, our thorough validation on multiple benchmarks demonstrates the effectiveness of SVTRv2. It achieves leading accuracy in various challenging scenarios covering regular, irregular, occluded, Chinese and long text, as well as whether employing pretraining. In addition, we retrain 24 methods from scratch on \textit{U14M-Filter} without data leakage. Their results on \textit{U14M} constitutes a comprehensive and reliable benchmark. We hope that SVTRv2 and this benchmark will further advance the development of the OCR community. 

\noindent\textbf{Acknowledgement}
This work was supported by the National Natural Science Foundation of China (Nos. 62427819, 32341012, 62172103).

{
    \small
    \bibliographystyle{ieeenat_fullname}
    \bibliography{main}
}

\clearpage
\setcounter{page}{1}
\maketitlesupplementary

\begin{table*}[t]\footnotesize
\centering
\setlength{\tabcolsep}{3.5  pt}{
\begin{tabular}{c|c|ccccccc|cccccccc|c|c|c|c}
\multicolumn{21}{c}{\setlength{\tabcolsep}{3pt}{\begin{tabular}{
>{\columncolor[HTML]{FFFFC7}}c 
>{\columncolor[HTML]{FFFFC7}}c 
>{\columncolor[HTML]{FFFFC7}}c 
>{\columncolor[HTML]{FFFFC7}}c 
>{\columncolor[HTML]{FFFFC7}}c 
>{\columncolor[HTML]{FFFFC7}}c c
>{\columncolor[HTML]{ECF4FF}}c 
>{\columncolor[HTML]{ECF4FF}}c 
>{\columncolor[HTML]{ECF4FF}}c 
>{\columncolor[HTML]{ECF4FF}}c 
>{\columncolor[HTML]{ECF4FF}}c 
>{\columncolor[HTML]{ECF4FF}}c 
>{\columncolor[HTML]{ECF4FF}}c }
\toprule
\textit{IIIT5k} & \textit{SVT} & \textit{ICDAR2013} & \textit{ICDAR2015} & \textit{SVTP} & \textit{CUTE80} & $\|$ & \textit{Curve} & \textit{Multi-Oriented} & \textit{Artistic} & \textit{Contextless} & \textit{Salient} & \textit{Multi-Words} & \textit{General} 
\end{tabular}}} \\
\toprule
   ID &  Method     & \multicolumn{6}{c}{\cellcolor[HTML]{FFFFC7}Common Benchmarks (\textit{Com})}   & Avg   & \multicolumn{7}{c}{\cellcolor[HTML]{ECF4FF}Union14M-Benchmark (\textit{U14M})}             & Avg & \textit{LTB}   & \textit{OST} & \textit{Size}  & \textit{FPS}   \\
\midrule
0 & SVTR (w/ TPS) &  98.1 &	96.1 &	96.4 &	89.2 &	92.1 &	95.8 &	94.62 &	82.2 &	86.1 &	69.7 &	75.1 &	81.6 &	73.8 &	80.7 &	78.44 & 0.0 & 71.2 & 19.95 &    141      \\
1 & 0 + w/o TPS & 98.0     & 97.1     & 97.3     & 88.6     & 90.7     & 95.8    & 94.58 & 76.2    & 44.5    & 67.8    & 78.7    & 75.2    & 77.9   & 77.8   & 71.17 & 45.1  & 67.8 & 18.10 &    161      \\
\midrule
2 & 1 + $\frac{H}{16}\rightarrow\frac{H}{8}$   & 98.9 &	97.4 &	97.9 &	89.7 &	91.8 &	96.9 &	95.41 &	82.2 &	64.3 &	70.2 &	80.0 &	80.9 &	80.6 &	80.5 &	76.95 & 44.8  &	69.5 & 18.10 &   145    \\
3 & 2 + Conv$^2$   & 98.7 & 97.1 & 97.1 & 89.6 & 91.6 & 97.6 & 95.28 & 82.9 & 65.6 & 73.2 & 80.0 & 80.5 & 81.6 & 80.8 & 77.78  & 47.4  & 71.1 & 17.77 &   159    \\
4 & 3 + MSR   & 98.7  & 98.0 & 97.4 & 89.4 & 91.6 & 97.6 & 95.44 & 87.4 & 83.7 & 75.4 & 80.9 & 81.9 & 83.5 & 82.8 & 82.22  & 50.9 & 72.5 & 17.77 &   159   \\
5 & 4 + FRM  & 98.8  & \textbf{98.1} & 98.4 & 89.8 & 92.9 & \textbf{99.0} & 96.16 & 88.2 & 86.2 & 77.5 & 83.2 & 83.9 & 84.6 & 83.5 & 83.86  & 50.7 & 74.9 & 19.76 &   143   \\
6 & 5 + SGM    & \textbf{99.2} & 98.0 & \textbf{98.7} & \textbf{91.1} & \textbf{93.5} & \textbf{99.0} & \textbf{96.57} & \textbf{90.6} & \textbf{89.0} & \textbf{79.3} & \textbf{86.1} & \textbf{86.2} & \textbf{86.7} & \textbf{85.1} & \textbf{86.14}   & \textbf{50.2}  & \textbf{80.0} & 19.76 &  143    \\
\bottomrule
\end{tabular}}
\caption{Ablation study of the proposed strategies on \textit{Com} and \textit{U14M}, along with their model sizes and FPS.}
\label{tab:add_ablation}
\end{table*}

\section{More Details of Ablation Study}

SVTRv2 builds upon the foundation of SVTR by introducing several innovative strategies aimed at addressing challenges in recognizing irregular text and modeling linguistic context. The key advancements and their impact are detailed as follows:

\textbf{Removal of the rectification Module and introduction of MSR and FRM}. In the original SVTR, a rectification module is employed to recognize irregular text. However, this approach negatively impacts the recognition of long text. To overcome this limitation, SVTRv2 removes the rectification module entirely. To effectively handle irregular text without compromising the CTC model's ability to generalize to long text, MSR and FRM are introduced.

\textbf{Improvement in feature resolution}. SVTR extracts visual representations of size \(\frac{H}{16} \times \frac{W}{4} \times D_2\) from input images of size \(H \times W \times 3\).  While this approach is effective for regular text, it struggles with retaining the distinct characteristics of irregular text. SVTRv2 doubles the height resolution ($\frac{H}{16}\rightarrow\frac{H}{8}$) of visual features, producing features of size \(\frac{H}{8} \times \frac{W}{4} \times D_2\), thereby improving its capacity to recognize irregular text.

\textbf{Refinement of local mixing mechanisms}. SVTR employs a hierarchical vision transformer structure, leveraging two mixing strategies: Local Mixing is implemented through a sliding window-based local attention mechanism, and Global Mixing employs the standard global multi-head self-attention mechanism. SVTRv2 retains the hierarchical vision transformer structure and the global multi-head self-attention mechanism for Global Mixing. For Local Mixing, SVTRv2 introduces a pivotal change. Specifically, the sliding window-based local attention is replaced with two consecutive group convolutions (Conv$^2$) \cite{he2016resnet}. It is important to highlight that unlike previous CNNs, there is no normalization or activation layer between the two convolutions.

\textbf{Semantic guidance module}. The original SVTR model relies solely on the CTC framework for both training and inference. However, CTC is inherently limited in its ability to model linguistic context. SVTRv2 addresses this by introducing a Semantic Guidance Module (SGM) during training. SGM facilitates the visual encoder in capturing linguistic information, enriching the feature representation. Importantly, SGM is discarded during inference, ensuring that the efficiency of CTC-based decoding remains unaffected while still benefiting from its contributions during the training phase.

\subsection{Progressive Ablation Experiments}

To comprehensively evaluate the contributions of every SVTRv2 upgrade, a series of progressive ablation experiments are conducted. Tab.~\ref{tab:add_ablation} outlines the results, along with the following observations:

1. Baseline (ID 0): The original SVTR serves as the baseline for comparison.

2. Rectification Module Removal (ID~1) reveals that while the rectification module (e.g., TPS) improves irregular text recognition accuracy, it hinders the model's ability to recognize long text. This confirms its limitations in balancing different recognition tasks.

3. Improvement in Feature Resolution (ID~2): Doubling the height resolution (\(\frac{H}{16} \rightarrow \frac{H}{8}\)) significantly boosts performance across challenging datasets, particularly for irregular text.

4. Replacement of Local Attention with Conv$^2$ (ID~3): Replacing the sliding window-based local attention with two consecutive group convolutions (Conv$^2$) yields improvements in artistic text, with a 3.0\% increase in accuracy. This result highlights the efficacy of convolution-based approaches in capturing character-level nuances, such as strokes and textures, thereby improving its ability to recognize artistic and irregular text.

5. Incorporation of MSR and FRM (ID~4 and ID~5): These components collectively enhance accuracy on irregular text benchmarks (e.g., \textit{Curve}), surpassing the rectification-based SVTR (ID~0) by 6.0\%, without compromising the CTC model's ability to generalize to long text.

6. Integration of SGM (ID~6): Adding SGM yields significant gains on multiple datasets, improving accuracy on \textit{OST} by 5.11\% and \textit{U14M} by 2.28\%.

It can be summarized as that, by integrating Conv$^2$, MSR, FRM, and SGM, SVTRv2 significantly improves performance in recognizing irregular text and modeling linguistic context over SVTR, while still maintaining robust long-text recognition capabilities and preserving the efficiency of CTC-based inference.

\section{SVTRv2 Variants}
There are several hyper-parameters in SVTRv2, including the depth of channel ($D_i$) and the number of heads at each stage, the number of mixing blocks ($N_i$) and their permutation. By varying them, SVTRv2 architectures with different capacities could be obtained and we construct three typical ones, i.e., SVTRv2-T (Tiny), SVTRv2-S (Small), SVTRv2-B (Base). Their detail configurations are shown in Tab.~\ref{tab:booktab1}.

In Tab.~\ref{tab:booktab1}, $[L]_m { [G]_n}$ denotes that the first \textit{m} mixing blocks in SVTRv2 utilize local mixing, while the last \textit{n} mixing blocks employ global mixing. Specifically, in SVTRv2-T and SVTRv2-S, all blocks in the first stage and the first three blocks in the second stage use local mixing. The last three blocks in the second stage, as well as all blocks in the third stage, are global mixing. In the case of SVTRv2-B, all blocks in the first stage and the first two blocks in the second stage use local mixing, whereas the last four blocks in the second stage and all blocks in the third stage adopt global mixing.

\begin{table}[t]\footnotesize
\centering
\setlength{\tabcolsep}{4pt}{
\begin{tabular}{c|c|c|c|c}
\toprule
Models & $\left[D_0, D_1, D_2 \right]$ & $[N_1, N_2, N_3]$ & Heads   & Permutation  \\
\midrule
SVTRv2-T & {[}64,128,256{]}  & {[}3,6,3{]}      & {[}2,4,8{]}  & $[L]_6 { [G]_6}$   \\
SVTRv2-S & {[}96,192,384{]}  & {[}3,6,3{]}      & {[}3,6,12{]}  & $[L]_6  {[G]_6 }$  \\
SVTRv2-B & {[}128,256,384{]} & {[}6,6,6{]}      & {[}4,8,12{]} & $[L]_8 { [G]_{10}}$ \\
\bottomrule
\end{tabular}}
\caption{Architecture specifications of SVTRv2 variants.}
\label{tab:booktab1}
\end{table}  

\begin{figure}[t]
  \centering
\includegraphics[width=0.47\textwidth]{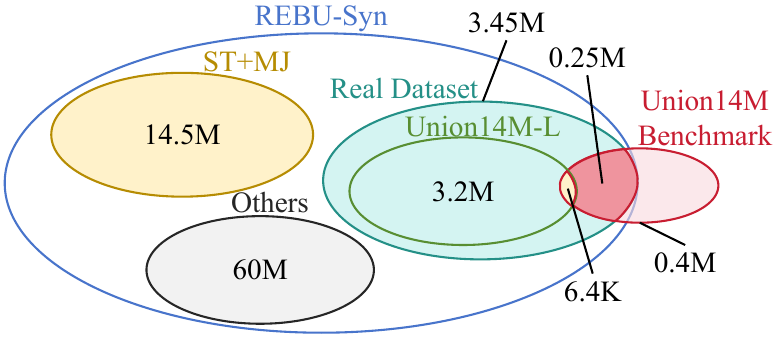} 
\caption{Relationships of the three real-world training sets and their overlapping with \textit{U14M}.}
\label{fig:data}
\end{figure}

\begin{table*}[t]\footnotesize
\centering
\setlength{\tabcolsep}{4pt}{
\begin{tabular}{c|ccccccc}
\toprule
& \textit{Curve} & \textit{Multi-Oriented} & \textit{Artistic} & \textit{Contextless} & \textit{Salient} & \textit{Multi-Words} & \textit{General} \\

& 2,426  & 1,369           & 900      & 779         & 1,585    & 829        & 400,000  \\
\midrule
\textit{Real} \cite{BautistaA22PARSeq}                & 1,276  & 440            & 432      & 326         & 431     & 193        & 254,174  \\
\textit{REBU-Syn} \cite{Rang_2024_CVPR_clip4str}            & 1,285  & 443            & 462      & 363         & 442     & 289        & 260,575 \\
\textit{U14M-Train} \cite{jiang2023revisiting}         & 9     & 3              & 30       & 37          & 11      & 96         & 6,401    \\

\bottomrule
\end{tabular}}
\caption{Overlapping statistics between three real-world training sets and \textit{U14M}.}
\label{tab:data}
\end{table*}

\begin{algorithm}[t]
\caption{Inference Time}
\label{alg:inferencetime}
\SetKwInOut{Input}{Input}\SetKwInOut{Output}{Output}
\Input{A set of images $\mathcal{I}$ with size $|\mathcal{I}| = 3000$, batch size $B = 1$, $N$ text lengths}
\Output{Overall inference time of the model}
\BlankLine
Initialize two lists: \texttt{total\_time\_list} and \texttt{count\_list} of size $N$, initialized to 0\;
\For{each image $I_j$ in $\mathcal{I}$ where $j \in \{1, 2, \ldots, 3000\}$}{
    Determine the text length $l_i$ for image $I_j$\;
    Perform inference on $I_j$ with text length $l_i$\;
    Record inference time $t_{ij}$\;
    \texttt{total\_time\_list[$l_i$]} += $t_{ij}$\;
    \texttt{count\_list[$l_i$]} += 1\;
}
\BlankLine
Initialize \texttt{avg\_time\_list}\;
\For{each text length $l_i$ where $i \in \{1, 2, \ldots, N\}$}{
    \If{\texttt{count\_list[$i$]} $> 0$}{
        \texttt{avg\_time\_list[$i$]} = \texttt{total\_time\_list[$i$]} / \texttt{count\_list[$i$]}\;
    }
}
\BlankLine
Compute the final average inference time:
\[
\texttt{inference\_time} = \frac{1}{N} \sum_{i=1}^{N} \texttt{avg\_time\_list[$i$]}
\]
\Return \texttt{inference\_time}\;
\end{algorithm}

\section{More Details of Real-World Datasets}

For English recognition, we train models on real-world datasets, from which the models exhibit stronger recognition capability \cite{BautistaA22PARSeq,jiang2023revisiting,Rang_2024_CVPR_clip4str}. There are three large-scale real-world training sets, i.e., the \textit{Real} dataset \cite{BautistaA22PARSeq}, \textit{REBU-Syn} \cite{Rang_2024_CVPR_clip4str}, and \textit{Union14M-L} (\textit{U14M-Train}) \cite{jiang2023revisiting}. However, as shown in Fig.~\ref{fig:data} and Tab.~\ref{tab:data}, the former two significantly overlap with \textit{U14M}, thus not suitable for model training when using \textit{U14M} at the evaluation dataset. Surprisingly, \textit{U14M-Train} is also overlapped with \textit{U14M} in nearly 6.5k text instances across the seven subsets. It means the models trained based on \textit{U14M-Train} suffer from data leakage when tested on \textit{U14M}, thus the results reported by \cite{jiang2023revisiting} should be updated. To this end, we create a filtered version of \textit{Union14M-L}, termed as \textit{U14M-Filter}, by filtering these overlapping instances from the training set. This new dataset is used to train SVTRv2 and other 24 methods we reproduced. 

\section{More Details of Inference Time}

In terms of the inference time, we do not utilize any acceleration framework and instead employ PyTorch's dynamic graph mode on one NVIDIA 1080Ti GPU. We first measure the inference time for 3,000 images with a batch size of \textit{1}, calculating the average inference time for each text length. We then compute the arithmetic mean of the average time across all text lengths to determine the overall inference time of the model. Algorithm~\ref{alg:inferencetime} details the process of measuring inference time.

\section{Results when Trained on Synthetic Datasets}

Previous research typically follows a typical evaluation protocol, where models are trained on synthetic datasets and validated using \textit{Com}, the six widely recognized real-world benchmarks. Following this protocol, we also train SVTRv2 and other models on synthetic datasets. In addition to evaluating SVTRv2 on \textit{Com}, we assess its performance on \textit{U14M}. The results offer a comprehensive evaluation of the model's generalization capabilities. For methods that have not reported performance on challenging benchmarks, we conduct additional evaluations using their publicly available models and present these results for comparative analysis. As illustrated in Tab.~\ref{tab:syn_sota}, models trained on synthetic datasets exhibit notably lower performance compared to those trained on large-scale real-world datasets (see Tab.~\ref{tab:sota}). This performance drop is particularly pronounced on challenging benchmarks. These findings highlight the critical importance of real-world datasets in improving recognition accuracy.

Despite trained on less diverse synthetic datasets, SVTRv2 also exhibits competitive performance. On irregular text benchmarks, such as \textit{Curve} and \textit{Multi-Oriented}, SVTR achieves strong results, largely due to its integrated rectification module \cite{shi2019aster}, which is particularly adept at handling irregular text patterns, even when trained on synthetic datasets. Notably, SVTRv2 achieves a substantial 4.8\% improvement over SVTR on \textit{Curve}, further demonstrating its enhanced capacity to address irregular text. Overall, these results demonstrate that, even when trained solely on synthetic datasets, SVTRv2 exhibits strong generalization capabilities, effectively handling complex and challenging text recognition scenarios.

\begin{table*}[t]\footnotesize
\centering
\setlength{\tabcolsep}{3pt}{
\begin{tabular}{r|c|c|ccccccc|cccccccc|c}
\multicolumn{19}{c}{\setlength{\tabcolsep}{3pt}{\begin{tabular}{
>{\columncolor[HTML]{FFFFC7}}c 
>{\columncolor[HTML]{FFFFC7}}c 
>{\columncolor[HTML]{FFFFC7}}c 
>{\columncolor[HTML]{FFFFC7}}c 
>{\columncolor[HTML]{FFFFC7}}c 
>{\columncolor[HTML]{FFFFC7}}c c
>{\columncolor[HTML]{ECF4FF}}c 
>{\columncolor[HTML]{ECF4FF}}c 
>{\columncolor[HTML]{ECF4FF}}c 
>{\columncolor[HTML]{ECF4FF}}c 
>{\columncolor[HTML]{ECF4FF}}c 
>{\columncolor[HTML]{ECF4FF}}c 
>{\columncolor[HTML]{ECF4FF}}c }
\toprule
\textit{IIIT5k} & \textit{SVT} & \textit{ICDAR2013} & \textit{ICDAR2015} & \textit{SVTP} & \textit{CUTE80} & $\|$ & \textit{Curve} & \textit{Multi-Oriented} & \textit{Artistic} & \textit{Contextless} & \textit{Salient} & \textit{Multi-Words} & \textit{General} 
\end{tabular}}} \\
\toprule
Method                         & Venue              & Encoder        & \multicolumn{6}{c}{\cellcolor[HTML]{FFFFC7}Common Benchmarks (\textit{Com})} & Avg   & \multicolumn{7}{c}{\cellcolor[HTML]{ECF4FF}Union14M-Benchmark (\textit{U14M})} & Avg  & \textit{Size} \\
\midrule
ASTER \cite{shi2019aster}        & TPAMI 2019 & ResNet+LSTM   & 93.3 & 90.0 & 90.8 & 74.7 & 80.2 & 80.9 & 84.98 & 34.0 & 10.2 & 27.7 & 33.0 & 48.2 & 27.6 & 39.8 & 31.50 & 27.2 \\
NRTR \cite{Sheng2019nrtr}         & ICDAR 2019 & Stem+TF$_6$       & 90.1 & 91.5 & 95.8 & 79.4 & 86.6 & 80.9 & 87.38 & 31.7 & 4.40  & 36.6 & 37.3 & 30.6 & 54.9 & 48.0 & 34.79 & 31.7 \\
MORAN \cite{pr2019MORAN}       & PR 2019    & ResNet+LSTM   & 91.0 & 83.9 & 91.3 & 68.4 & 73.3 & 75.7 & 80.60 & 8.90  & 0.70  & 29.4 & 20.7 & 17.9 & 23.8 & 35.2 & 19.51 & 17.4     \\
SAR \cite{li2019sar}         & AAAI 2019  & ResNet+LSTM   & 91.5 & 84.5 & 91.0 & 69.2 & 76.4 & 83.5 & 82.68 & 44.3 & 7.70  & 42.6 & 44.2 & 44.0 & 51.2 & 50.5 & 40.64 & 57.7 \\
DAN \cite{wang2020aaai_dan}        & AAAI 2020  & ResNet+FPN    & 93.4 & 87.5 & 92.1 & 71.6 & 78.0 & 81.3 & 83.98 & 26.7 & 1.50  & 35.0 & 40.3 & 36.5 & 42.2 & 42.1 & 32.04 & 27.7     \\
SRN \cite{yu2020srn}         & CVPR 2020  & ResNet+FPN    & 94.8 & 91.5 & 95.5 & 82.7 & 85.1 & 87.8 & 89.57 & 63.4 & 25.3 & 34.1 & 28.7 & 56.5 & 26.7 & 46.3 & 40.14 & 54.7 \\
SEED* \cite{cvpr2020seed}        & CVPR 2020  & ResNet+LSTM   &  93.8    &  89.6    & 92.8     & 80.0     & 81.4     & 83.6     &  86.87     &  40.4    &  15.5    &  32.1    &    32.5  &  54.8    &  35.6    & 39.0     &  35.70     &  24.0    \\
AutoSTR* \cite{zhang2020autostr}     & ECCV 2020  & NAS+LSTM      & 94.7     & 90.9     & 94.2     &  81.8    &  81.7    &  -    &   -    &  47.7    &  17.9    & 30.8     &   36.2   & 64.2     & 38.7     &  41.3    & 39.54      &  6.00    \\
RoScanner \cite{yue2020robustscanner}   & ECCV 2020  & ResNet        & 95.3 & 88.1 & 94.8 & 77.1 & 79.5 & 90.3 & 87.52 & 43.6 & 7.90  & 41.2 & 42.6 & 44.9 & 46.9 & 39.5 & 38.09 & 48.0 \\
ABINet \cite{fang2021abinet}       & CVPR 2021  & ResNet+TF$_3$     & 96.2 & 93.5 & 97.4 & 86.0 & 89.3 & 89.2 & 91.93 & 59.5 & 12.7 & 43.3 & 38.3 & 62.0 & 50.8 & 55.6 & 46.03 & 36.7 \\
VisionLAN \cite{Wang_2021_visionlan}   & ICCV 2021  & ResNet+TF$_3$     & 95.8 & 91.7 & 95.7 & 83.7 & 86.0 & 88.5 & 90.23 & 57.7 & 14.2 & 47.8 & 48.0 & 64.0 & 47.9 & 52.1 & 47.39 & 32.8 \\
PARSeq* \cite{BautistaA22PARSeq}     & ECCV 2022  & ViT-S         & 97.0 & 93.6 & 97.0 & 86.5 & 88.9 & 92.2 & 92.53 & 63.9 & 16.7 & 52.5 & 54.3 & 68.2 & 55.9 & 56.9 & 52.62 & 23.8 \\
MATRN \cite{MATRN}       & ECCV 2022  & ResNet+TF$_3$     & 96.6 & 95.0 & 97.9 & 86.6 & 90.6 & 93.5 & 93.37 & 63.1 & 13.4 & 43.8 & 41.9 & 66.4 & 53.2 & 57.0 & 48.40 & 44.2 \\
MGP-STR* \cite{mgpstr}    & ECCV 2022  & ViT-B         & 96.4 & 94.7 & 97.3 & 87.2 & 91.0 & 90.3 & 92.82 & 55.2 & 14.0 & 52.8 & 48.5 & 65.2 & 48.8 & 59.1 & 49.09 & 148  \\
LevOCR* \cite{levocr}     & ECCV 2022  &    ResNet+TF$_3$            & 96.6 & 94.4 & 96.7 & 86.5 & 88.8 & 90.6 & 92.27 & 52.8 & 10.7 & 44.8 & 51.9 & 61.3 & 54.0 & 58.1 & 47.66 & 109  \\
CornerTF* \cite{xie2022toward_cornertrans} & ECCV 2022  &    CornerEncoder           & 95.9 & 94.6 & 97.8 & 86.5 & 91.5 & 92.0 & 93.05 & 62.9 & 18.6 & 56.1 & 58.5 & 68.6 & 59.7 & 61.0 & 55.07 & 86.0 \\
SIGA* \cite{Guan_2023_CVPR_SIGA}       & CVPR 2023  &     ViT-B          & 96.6 & 95.1 & 97.8 & 86.6 & 90.5 & 93.1 & 93.28 & 59.9 & 22.3 & 49.0 & 50.8 & 66.4 & 58.4 & 56.2 & 51.85 & 113  \\
CCD* \cite{Guan_2023_ICCV_CCD}        & ICCV 2023  &      ViT-B         & 97.2 & 94.4 & 97.0 & 87.6 & 91.8 & 93.3 & 93.55 & 66.6 & 24.2 & \textbf{63.9} & 64.8 & 74.8 & 62.4 & 64.0 & 60.10 & 52.0 \\
LISTER* \cite{iccv2023lister}     & ICCV 2023  & FocalNet-B    & 96.9 & 93.8 & 97.9 & 87.5 & 89.6 & 90.6 & 92.72 & 56.5 & 17.2 & 52.8 & 63.5 & 63.2 & 59.6 & 65.4 & 54.05 & 49.9 \\
LPV-B* \cite{ijcai2023LPV}      & IJCAI 2023 & SVTR-B        & 97.3 & 94.6 & 97.6 & 87.5 & 90.9 & 94.8 & 93.78 & 68.3 & 21.0 & 59.6 & 65.1 & 76.2 & 63.6 & 62.0 & 59.40 & 35.1 \\
CDistNet* \cite{zheng2024cdistnet} & IJCV 2024  & ResNet+TF$_3$     & 96.4 & 93.5 & 97.4 & 86.0 & 88.7 & 93.4 & 92.57 & 69.3 & 24.4 & 49.8 & 55.6 & 72.8 & 64.3 & 58.5 & 56.38 & 65.5 \\
CAM* \cite{yang2024class_cam}        & PR 2024    & ConvNeXtV2-B    & 97.4 & \textbf{96.1} & 97.2 & 87.8 & 90.6 & 92.4 & 93.58 & 63.1 & 19.4 & 55.4 & 58.5 & 72.7 & 51.4 & 57.4 & 53.99 & 135  \\
BUSNet \cite{Wei_2024_busnet}      & AAAI 2024  & ViT-S         & 96.2 & 95.5 & 98.3 & 87.2 & 91.8 & 91.3 & 93.38 & -    & -    & -    & -    & -    & -    & -    & -     & 56.8 \\
DCTC \cite{Zhang_Lu_Liao_Huang_Li_Wang_Peng_2024_DCTC}        & AAAI 2024  & SVTR-L        & 96.9 & 93.7 & 97.4 & 87.3 & 88.5 & 92.3 & 92.68 & -    & -    & -    & -    & -    & -    & -    & -     & 40.8 \\
OTE \cite{Xu_2024_CVPR_OTE}         & CVPR 2024  & SVTR-B        & 96.4 & 95.5 & 97.4 & 87.2 & 89.6 & 92.4 & 93.08 & -    & -    & -    & -    & -    & -    & -    & -     & 25.2 \\
CPPD \cite{du2023cppd}        & TPAMI 2025  & SVTR-B        & 97.6 & 95.5 & 98.2 & 87.9 & 90.9 & 92.7 & 93.80 & 65.5 & 18.6 & 56.0 & 61.9 & 71.0 & 57.5 & 65.8 & 56.63 & 26.8 \\
IGTR-AR \cite{du2024igtr} & TPAMI 2025  & SVTR-B       & \textbf{98.2}     &  95.7    &   \textbf{98.6}   &    \textbf{88.4}  &   \textbf{92.4}   &  95.5    &  \textbf{94.78}                                                                           &  \textbf{78.4}     &    31.9                                                       &      61.3    &   66.5          &   \textbf{80.2}      &    69.3                                                    &  \textbf{67.9}       &    \textbf{65.07}                                                                         &   24.1
                         \\
SMTR \cite{du2024smtr} & AAAI 2025  & FocalSVTR& 97.4 & 94.9 & 97.4 & \textbf{88.4} & 89.9 & \textbf{96.2} & 94.02 & 74.2 & 30.6 & 58.5 & 67.6 & 79.6 & \textbf{75.1} & \textbf{67.9} & 64.79  &  15.8 \\
\midrule
CRNN \cite{shi2017crnn}        & TPAMI2016 & ResNet+LSTM & 82.9 & 81.6 & 91.1 & 69.4 & 70.0 & 65.5 & 76.75 & 7.50  & 0.90  & 20.7 & 25.6 & 13.9 & 25.6 & 32.0 & 18.03 & 8.30 \\
SVTR* \cite{duijcai2022svtr}     & IJCAI2022 & SVTR-B        & 96.0 & 91.5 & 97.1 & 85.2 & 89.9 & 91.7 & 91.90 & 69.8 & \textbf{37.7} & 47.9 & 61.4 & 66.8 & 44.8 & 61.0 & 55.63 & 24.6 \\
\cellcolor[HTML]{EFEFEF}SVTRv2       &    \cellcolor[HTML]{EFEFEF}-      & \cellcolor[HTML]{EFEFEF}SVTRv2-B      & \cellcolor[HTML]{EFEFEF}97.7 & \cellcolor[HTML]{EFEFEF}94.0 & \cellcolor[HTML]{EFEFEF}97.3 & \cellcolor[HTML]{EFEFEF}88.1 & \cellcolor[HTML]{EFEFEF}91.2 & \cellcolor[HTML]{EFEFEF}95.8 & \cellcolor[HTML]{EFEFEF}94.02 & \cellcolor[HTML]{EFEFEF}74.6 & \cellcolor[HTML]{EFEFEF}25.2 & \cellcolor[HTML]{EFEFEF}57.6 & \cellcolor[HTML]{EFEFEF}\textbf{69.7} & \cellcolor[HTML]{EFEFEF}77.9 & \cellcolor[HTML]{EFEFEF}68.0 & \cellcolor[HTML]{EFEFEF}66.9 & \cellcolor[HTML]{EFEFEF}62.83 & \cellcolor[HTML]{EFEFEF}19.8 \\
\bottomrule
\end{tabular}}
\caption{Results of SVTRv2 and existing models when trained on synthetic datasets (\textit{ST} + \textit{MJ}) \cite{Synthetic,jaderberg14synthetic}. * represents that the results on \textit{U14M} are evaluated using the model they released.}
\label{tab:syn_sota}
\end{table*}

\section{Qualitative Analysis of Recognition Results}

\begin{figure*}[t]
  \centering
\includegraphics[width=0.98\textwidth]{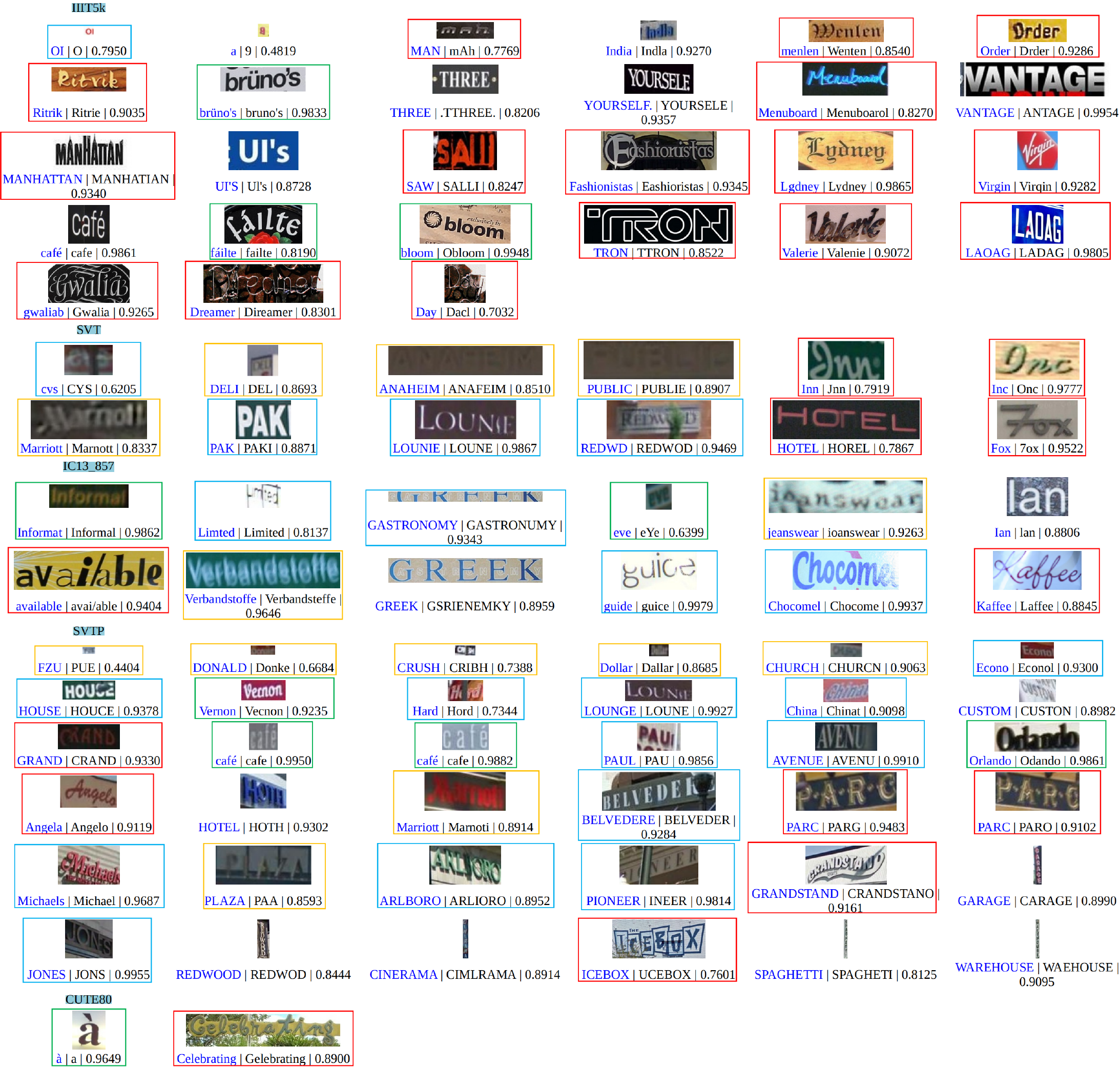} 
\caption{The bad cases of SVTRv2 in IIIT5k \cite{IIIT5K}, SVT \cite{Wang2011SVT}, ICDAR 2013 \cite{icdar2013}, SVTP \cite{SVTP} and CUTE80 \cite{Risnumawan2014cute}.  Labels, the predicted result, and the predicted score are denoted as \textcolor{blue}{Text$_{label}$} $|$ Text$_{pred}~|~$ Score$_{pred}$. Yellow, red, blue, and green boxes indicate blurred, artistic fonts, incomplete text, and label-inconsistent samples, respectively. Other samples have no box.}
\label{fig:noic15}
\end{figure*}

\begin{figure*}[t]
  \centering
\includegraphics[width=0.98\textwidth]{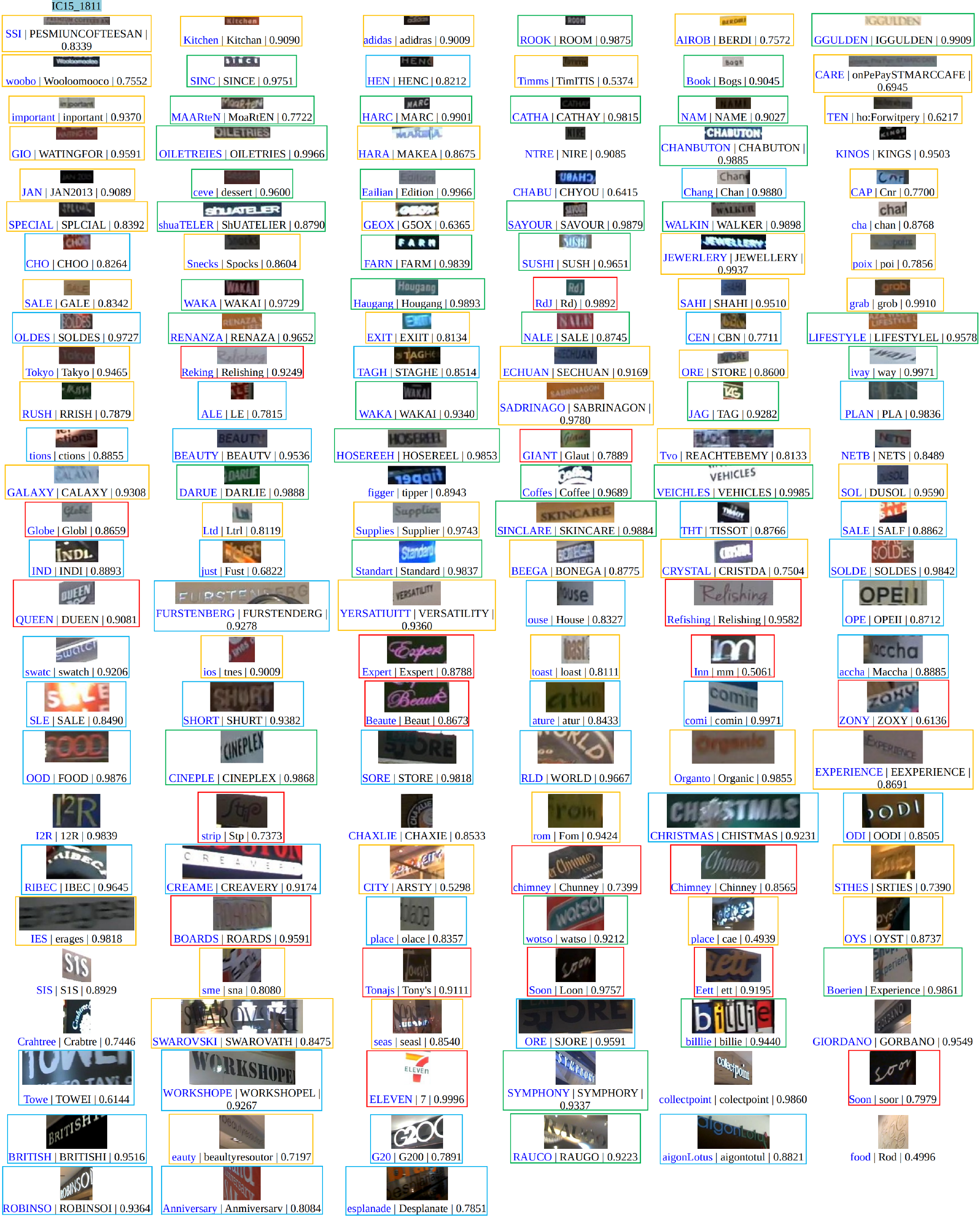} 
\caption{The bad cases of SVTRv2 in ICDAR 2015 \cite{icdar2015}. Labels, the predicted result, and the predicted score are denoted as \textcolor{blue}{Text$_{label}$} $|$ Text$_{pred}~|~$ Score$_{pred}$. Yellow, red, blue, and green boxes indicate blurred, artistic fonts, incomplete text, and label-inconsistent samples, respectively. Other samples have no box.}
\label{fig:ic15}
\end{figure*}

The SVTRv2 model achieved an average accuracy of 96.57\% on \textit{Com} (see Tab. \ref{tab:sota}). To investigate the underlying causes of the remaining 3.43\% of recognition errors, we conducted a detailed analysis of the misclassified samples, as illustrated in Fig.~\ref{fig:noic15} and Fig.~\ref{fig:ic15}. While previous research has typically categorized \textit{Com} into \textit{regular} and \textit{irregular} text. However, these error samples indicate that the majority of incorrectly recognized text is not irregular. This suggests that, under the current training paradigm using large-scale real-world datasets, a more rigorous manual screening process is warranted for common benchmarks.

\begin{table}[t]\footnotesize
\centering
\setlength{\tabcolsep}{1pt}{\begin{tabular}{c|cccc|c|c}
\toprule
           & Blurred & Artistic & Incomplete & Other   & Total & Label$_{err}$ \\
\midrule
IIIT5k \cite{IIIT5K}     & 0        & 16       & 1        & 4       & 21    & 4     \\
SVT \cite{Wang2011SVT}       & 4        & 4        & 4      & 0       & 12    & 0     \\
ICDAR 2013 \cite{icdar2013} & 2        & 2        & 4       & 2       & 10    & 2     \\
ICDAR 2015 \cite{icdar2015}  & 48       & 19       & 42       & 13      & 122   & 35    \\
SVTP \cite{SVTP}       & 7        & 6        & 12       & 7       & 32    & 4     \\
CUTE80 \cite{Risnumawan2014cute}     & 0        & 1        & 0         & 0       & 1     & 1     \\
\midrule
Total      & 61       & 48       & 63        & 26      & 198   & 46    \\
           & 30.81\%  & 24.24\%  & 31.82\%   & 13.13\% & 100\% &     \\
\bottomrule
\end{tabular}}
\caption{Distribution of bad cases for SVTRv2 on \textit{Com}.}
\label{tab:bad_case}
\end{table}

Based on this one-by-one manual viewing, we identified five primary causes of recognition errors: (1) blurred, (2) artistic, (3) incomplete text, (4) others, and (5) image text labeling errors (Label$_{err}$). Specifically, the blurring text includes issues such as low resolution, motion blur, or extreme lighting conditions. The artistic text category refers to unconventional fonts, commonly found in business signage, as well as some handwritten text. Incomplete text arises when characters are obscured by objects or lost due to improper cropping, requiring contextual inference. Image text labeling errors occur when the given text labels contain inaccuracies or include characters with phonetic symbols. As shown in Tab.~\ref{tab:bad_case}, after excluding samples affected by labeling inconsistencies, the remaining recognition errors primarily stemmed from blurred (30.81\%), artistic (24.24\%), and incomplete text (31.82\%). This result highlights that SVTRv2's recognition performance needs further improvement, particularly in handling complex scenarios involving these challenging text types.

\section{Standardized Model Training Settings}

The optimal hyperparameters for training different models vary and are not universally fixed. However, key factors such as training epochs, data augmentations, input size, and evaluation protocols significantly influence model accuracy. To ensure fair and unbiased performance comparisons, we standardize these factors across all models, as outlined in Tab.~\ref{tab:setting}. This uniform training and evaluation framework ensures consistency while allowing each model to approach its best accuracy. To maximize fairness, we conducted extensive hyperparameter tuning for model-specific settings, including the optimizer, learning rate, and regularization strategies. This rigorous optimization led to significant accuracy improvements of 5–10\% for most models compared to their default configurations. For instance, MAERec’s accuracy increased from 78.6\% to 85.2\%, demonstrating the effectiveness of training settings. These improvements underscore the reliability of our results and highlight the importance of carefully optimizing hyperparameters for meaningful model comparisons.

\begin{table*}
\centering
\begin{tabular}{c|p{10cm}}
\toprule
\textbf{Setting} & \textbf{Detail} \\
\midrule
\textbf{Training Set} &
For training, when the text length of a text image exceeds 25, samples with text length $\leq 25$ are randomly selected from the training set to ensure models are only exposed to short texts (length $\leq 25$). \\
\midrule
\textbf{Test Sets} &
For all test sets except the long-text test set (\textit{LTB}), text images with text length $> 25$ are filtered. Text length is calculated by removing spaces and non-94-character-set special characters. \\
\midrule
\textbf{Input Size} &
Unless a method explicitly requires a dynamic size, models use a fixed input size of $32\times128$. If a model performs incorrectly with $32\times128$ during training, the original size is used. The test input size matches the training size. \\
\midrule
\textbf{Data Augmentation} &
All models use the data augmentation strategy employed by PARSeq. \\
\midrule
\textbf{Training Epochs} &
Unless pre-training is required, all models are trained for 20 epochs. \\
\midrule
\textbf{Optimizer} &
AdamW is the default optimizer. If training fails to converge with AdamW, Adam or other optimizers are used. \\
\midrule
\textbf{Batch Size} &
Maximum batch size for all models is 1024. If single-GPU training is not feasible, 2 GPUs (512 per GPU) or 4 GPUs (256 per GPU) are used. If 4-GPU training runs out of memory, the batch size is halved, and the learning rate is adjusted accordingly. \\
\midrule
\textbf{Learning Rate} &
Default learning rate for batch size 1024 is 0.00065. The learning rate is adjusted multiple times to achieve the best results. \\
\midrule
\textbf{Learning Rate Scheduler} &
A linear warm-up for 1.5 epochs is followed by a OneCycle scheduler. \\
\midrule
\textbf{Weight Decay} &
Default weight decay is 0.05. NormLayer and Bias parameters have a weight decay of 0. \\
\midrule
\textbf{EMA or Similar Tricks} &
No EMA or similar tricks are used for any model. \\
\midrule
\textbf{Evaluation Protocols} &
Word accuracy is evaluated after filtering special characters and converting all text to lowercase. \\
\bottomrule
\end{tabular}

\caption{A uniform training and evaluation setting to maintain consistency across all settings while simultaneously enabling each model to achieve its best possible accuracy.}
\label{tab:setting}
\end{table*}

\end{document}